\documentclass[10pt, letter, onecolumn]{arxiv}

\usepackage{helvet}
\usepackage{kantlipsum, lipsum}
\usepackage{dm-colors}
\usepackage{amsmath}
\usepackage{pstricks, pst-node}
\usepackage{verbatim}
\usepackage{multirow}
\usepackage{array}
\usepackage{longtable}
\usepackage{pdflscape}
\usepackage{scalerel}
\usepackage{booktabs}
\usepackage{enumitem}
\usepackage{xspace}
\usepackage{bm}
\usepackage{bbm}
\usepackage{mathtools}
\usepackage{soul}
\usepackage{epsfig}
\usepackage{graphicx}
\usepackage{amssymb}
\usepackage{colortbl}
\usepackage{csquotes}
\usepackage{setspace}
\usepackage{bbding}
\usepackage{siunitx} 
\usepackage{threeparttable}
\usepackage{tabularx,ragged2e}
\usepackage{placeins}
\usepackage{bbding}
\definecolor{darkpastelgreen}{rgb}{0.13, 0.55, 0.13}
\definecolor{darkpastelred}{rgb}{0.55, 0.13, 0.13}
\definecolor{mygray}{rgb}{1, 1, 1}
\usepackage{lmodern}
\usepackage[hang,flushmargin]{footmisc}

\usepackage{nameref}
\usepackage{varioref}
\usepackage{amssymb}
\usepackage{pifont}
\usepackage{rotating}
\usepackage{graphicx}

\usepackage[pagebackref=false,breaklinks=false,%
            colorlinks=true,bookmarks=true,citecolor=ourdarkblue,%
            urlcolor=ourdarkblue,linkcolor=ourdarkblue]{hyperref}
\usepackage[noabbrev,capitalize]{cleveref}
\usepackage{etoc}
\usepackage{lineno}
\usepackage{tikz}

\usepackage{wasysym}
\usepackage{algorithm}
\usepackage{makecell} 
\usepackage{algpseudocode}

\usepackage{amsmath}
\usepackage{thmtools}
\usepackage{tcolorbox}
\usepackage{wasysym}
\declaretheoremstyle[
    spaceabove=6pt, spacebelow=6pt,
    headfont=\bfseries, headpunct={.}, headformat={\NAME\ \NUMBER},
    bodyfont=\normalfont,
    postheadspace=0.5em
]{promptstyle}

\tcolorboxenvironment{prompt}{
    colback=gray!10!,
    colframe=gray!75!,
    fonttitle=\bfseries,
    title=Prompt,
    boxrule=0.5pt,
    sharp corners
}

\graphicspath{{figures/}}
\definecolor{mygray}{rgb}{0.85, 0.85, 0.85}

\usepackage{listings}
\usepackage{xcolor}

\definecolor{codegreen}{rgb}{0,0.6,0}
\definecolor{codegray}{rgb}{0.5,0.5,0.5}
\definecolor{codepurple}{rgb}{0.58,0,0.82}
\definecolor{backcolour}{rgb}{0.95,0.95,0.92}
\definecolor{framecolor}{rgb}{0.8,0.8,0.8}

\lstdefinestyle{prettyjson}{
    backgroundcolor=\color{backcolour},   
    commentstyle=\color{codegreen},
    keywordstyle=\color{blue}\bfseries,
    numberstyle=\tiny\color{codegray},
    stringstyle=\color{codepurple},
    basicstyle=\ttfamily\small,
    breakatwhitespace=false,         
    breaklines=true,                 
    captionpos=b,                    
    keepspaces=true,                 
    numbers=left,                    
    numbersep=8pt,                  
    showspaces=false,                
    showstringspaces=false,
    showtabs=false,                  
    tabsize=2,
    frame=single,
    frameround=tttt,
    framerule=0.5pt,
    rulecolor=\color{framecolor},
    xleftmargin=15pt,
    xrightmargin=15pt,
    aboveskip=15pt,
    belowskip=15pt,
    columns=flexible,
    escapeinside={(*@}{@*)}
}

\title{\Large{DentVLM: A Multimodal Vision-Language Model for Comprehensive Dental Diagnosis and Enhanced Clinical Practice}}

\author[1,2,$\ast$]{Zijie Meng} 
\author[3,$\ast$,$\dagger$]{Jin Hao} 
\author[1,2,$\ast$]{Xiwei Dai}
\author[4]{Yang Feng}
\author[2]{Jiaxiang Liu}
\author[1]{Bin Feng}
\author[4]{Huikai Wu}
\author[1,2]{Xiaotang Gai}
\author[1,2]{Hengchuan Zhu}
\author[1,2]{Tianxiang Hu}
\author[2]{Yangyang Wu}
\author[5]{Hongxia Xu}
\author[6]{Jin Li}
\author[2]{Jun Xiao}
\author[7]{Xiaoqiang Liu}
\author[8]{Joey Tianyi Zhou}
\author[1]{Fudong Zhu}
\author[9]{Zhihe Zhao}
\author[3]{Lunguo Xia}
\author[3]{Bing Fang}
\author[10,$\dagger$]{Jimeng Sun}
\author[2,5,$\dagger$]{Jian Wu}
\author[1,2,5,$\dagger$]{Zuozhu Liu}

\affil[1]{\normalsize Stomatology Hospital, School of Stomatology, Zhejiang University School of Medicine, Zhejiang University, Hangzhou 310016, Zhejiang, China. \authorcr \vspace{0.1cm}}

\affil[2]{\normalsize College of Computer Science and Technology, Zhejiang University-University of Illinois Urbana-Champaign Institute, Zhejiang University, Hangzhou 310027, Zhejiang, China. \authorcr \vspace{0.1cm}
}

\affil[3]{\normalsize Department of Orthodontics, Shanghai Ninth People's Hospital, College of Stomatology, Shanghai Jiao Tong University School of Medicine, Shanghai 200011, China. \authorcr \vspace{0.1cm}
}

\affil[4]{\normalsize Angelalign Technology Inc., Shanghai 200082, China \authorcr \vspace{0.1cm}
}

\affil[5]{\normalsize Zhejiang Key Laboratory of Medical Imaging Artificial Intelligence, Haining 314400, Zhejiang, China \authorcr \vspace{0.1cm}
}

\affil[6]{\normalsize Department of Stomatology, The First Affiliated Hospital of Shenzhen University, Shenzhen Second People's Hospital, Shenzhen 518035, China \authorcr \vspace{0.1cm}
}

\affil[7]{\normalsize Department of Prosthodontics, Peking University School and Hospital of Stomatology, Beijing 100081, China \authorcr \vspace{0.1cm}
}

\affil[8]{\normalsize CFAR \& IHPC, Agency for Science, Technology and Research, 138632, Singapore \authorcr \vspace{0.1cm}
}

\affil[9]{\normalsize State Key Laboratory of Oral Diseases, National Clinical Research Center for Oral Diseases, West China Hospital of Stomatology, Sichuan University, Chengdu, China \authorcr \vspace{0.1cm}
}

\affil[10]{\normalsize Siebel School of Computing and Data Science, University of Illinois Urbana-Champaign, Urbana, IL 61801, USA \authorcr \vspace{0.1cm}
}

\affil[$\ast$]{\normalsize Equal contributions\hspace{1cm}}
\affil[$\dag$]{\normalsize Corresponding author\authorcr haojin@stanford.edu (J.H.); jimeng@illinois.edu (J.S.); wujian2000@zju.edu.cn (J.W.); zuozhuliu@intl.zju.edu.cn (Z.L.)}

\begin{document}

\begin{abstract}
Diagnosing and managing oral diseases necessitate advanced visual interpretation across diverse imaging modalities and integrated information synthesis. While current AI models excel at isolated tasks, they often fall short in addressing the complex, multimodal requirements of comprehensive clinical dental practice. Here we introduce DentVLM, a multimodal vision-language model engineered for expert-level oral disease diagnosis. DentVLM was developed using a comprehensive, large-scale, bilingual dataset of 110,447 images and 2.46 million visual question-answering (VQA) pairs. The model is capable of interpreting seven 2D oral imaging modalities across 36 diagnostic tasks, significantly outperforming leading proprietary and open-source models by 19.6\% higher accuracy for oral diseases and 27.9\% for malocclusions. In a clinical study involving 25 dentists, evaluating 1,946 patients and encompassing 3,105 QA pairs, DentVLM surpassed the diagnostic performance of 13 junior dentists on 21 of 36 tasks and exceeded that of 12 senior dentists on 12 of 36 tasks. When integrated into a collaborative workflow, DentVLM elevated junior dentists' performance to senior levels and reduced diagnostic time for all practitioners by 15-22\%. Furthermore, DentVLM exhibited promising performance across three practical utility scenarios, including home-based dental health management, hospital-based intelligent diagnosis and multi-agent collaborative interaction. These findings establish DentVLM as a robust clinical decision support tool, poised to enhance primary dental care, mitigate provider-patient imbalances, and democratize access to specialized medical expertise within the field of dentistry.

\end{abstract}

\maketitle


\section{Introduction}

Oral diseases constitute a formidable public health challenge, affecting approximately 3.5 billion people globally, with a particularly acute burden in low- and middle-income countries (LMICs)~\cite{jain2024s}. The disparity in professional access, a dentist-to-population ratio of approximately 1:360,000 in these regions compared to 1:2,000 in high-income countries, underscores a critical global health inequity~\cite{jain2024s,nyamuryekung2015relative,uguru2020access}. This gap leads to substantial morbidity, as preventable conditions like chronic dental pain compromise overall well-being, educational attainment, and workforce productivity~\cite{masood2015household}. Access to specialized services such as orthodontics is even more limited due to specialist shortages and insufficient public health awareness~\cite{boutayeb2010burden,seminario2020mitigating,luan2024universal}. The combination of increasing disease prevalence and constrained healthcare capacity creates a cycle of delayed diagnoses, suboptimal management, and ineffective referral pathways, necessitating immediate and scalable interventions~\cite{benzian2011political,watt2019ending}.

Conventional oral health assessments, which depend on clinical inspection and radiographic evaluation, are often inefficient and prone to variability between practitioners. These limitations are pronounced in settings with limited resources. While emerging digital technologies like teledentistry, AI-powered radiographic analysis, and portable imaging devices aim to enhance clinical decision-making, they largely function in fragmented silos~\cite{rashid2022hybrid,islam2022teledentistry,jang2022accurate,cui2019toothnet,liu2023deep,hao2022toward,xiong2023tsegformer,shi2024leta,elgarba2024validation,abdinian2024comparison,park2009portable}. These solutions typically address either diagnostics or care delivery optimization in isolation, failing to offer a comprehensive, integrated framework. Effective dental practice is inherently multimodal, requiring the synthesis of diverse data from clinical photographs and radiographs to patient histories. The utility of digital tools is further restricted by their reliance on clinical expertise, which remains scarce in areas where it is most needed~\cite{glick2021fdi}. 

Recent breakthroughs in foundation models and Vision-Language Models (VLMs) offer a new paradigm for reshaping healthcare delivery through advanced multimodal understanding and reasoning capabilities~\cite{patel2023chatgpt,ali2023using,singhal2023large}. Although these models show promise in addressing general medical queries~\cite{singhal2023large,ayers2023comparing}, their application to specialized domains like dentistry remains nascent~\cite{umer2024innovation,huang2023chatgpt}. Adapting general-purpose models to the intricacies of clinical dentistry necessitates overcoming challenges related to multimodal data fusion and embedding domain-specific knowledge to interpret highly specific dental imaging modalities~\cite{huang2023chatgpt}. 

To address these systemic deficiencies, we developed the \textbf{Dent}al \textbf{V}ision-\textbf{L}anguage \textbf{M}odel (DentVLM), a versatile VLM engineered to transform oral disease diagnostics and management. DentVLM is designed to interpret seven distinct 2D oral imaging modalities and perform 36 diagnostic tasks related to a wide spectrum of oral diseases and malocclusions. To empower the model with deep domain expertise, we curated a comprehensive, large-scale, bilingual dataset of 110,447 images and over 2.4 million visual question-answering (VQA) pairs. Utilizing this dataset, we implemented a two-stage training strategy: initial vision-language alignment followed by diagnosis instruction tuning, adapting the foundational Qwen2-VL-7B model for the complexities of clinical dentistry. 

We systematically validated DentVLM's performance against established baselines and through a comprehensive clinical reader study. It demonstrated superior performance in direct comparisons with 18 state-of-the-art models~\cite{gpt4o,gemini,deepseekvl,gemma3,internvl2.5,llava-onevision,llama3,qwen2vl,qwen2.5vl,llavamed,huatuogptvision,radfm}, including proprietary, open-source generic, and medical VLMs. DentVLM substantially outperformed the next-leading models in diagnostic accuracy of oral diseases and malocclusions by 19.61\% and 27.91\% (all with $P<0.001$), respectively. In a clinical study with 25 dentists, DentVLM’s diagnostic accuracy surpassed 13 junior clinicians in 21 of 36 tasks and 12 senior ones in 12 of 36 tasks. Moreover, in a human-AI collaborative setting, DentVLM significantly enhanced the diagnostic capabilities of both junior and senior dentists. It elevated junior dentists' accuracy to levels comparable with senior practitioners while reducing diagnostic time by 15-22\% for all (all with $P<0.001$). Additionally, DentVLM's practical utility was explored in health management, intelligent diagnosis, and multi-agent collaborative interaction, indicating its broad and promising potential for real-world applications.

Our findings establish DentVLM as a robust clinical decision support tool with the potential to enhance the quality of primary dental care, mitigate provider-patient imbalances, and democratize access to specialized medical expertise. This work provides a framework for developing specialty-specific foundation models that integrate complex imaging and clinical workflows, offering a scalable approach to addressing pressing global health challenges.

\section{Results}
\subsection{Problem formulation and dataset}
We engineered DentVLM as a versatile vision-language model, designed to diagnose 36 distinct dental conditions across seven diverse imaging modalities. Its analytical capabilities extend to both radiographic inputs, specifically panoramic (PAN) and lateral (LAT) X-rays, and a comprehensive suite of intraoral photographs, including frontal (INF), left (INL), right (INR), upper arch (UPP), and lower arch (LOW) views (Extended Table 1, Supplementary Note 1). The model operates within a predefined matrix of diagnostic tasks, where each condition is mapped to specific imaging modalities based on a consensus framework developed by a panel of senior dental experts (Figure~\ref{fig:figure1}.a, Extended Table 2-4, Supplementary Note 1). These 36 conditions are categorized into two domains: oral diseases and malocclusions. 

The diagnostic process within DentVLM is operated through a VQA paradigm. Specifically, for an input pair consisting of an image $I_t$ and question $Q_t$ corresponding to task $t$, denoted as $(I_t, Q_t)$, DentVLM generates a comprehensive output that includes an answer $A_t$, a detailed rationale $R_t$, and, where applicable, a precise anatomical location $L_t$. This can be formally expressed as $\text{DentVLM}(I_t, Q_t)=(A_t, R_t, L_t)$, with $L_t$ being optional for malocclusion diagnosis (Figure~\ref{fig:figure1}.a, Extended Table 5, Supplementary Note 1). While the majority of these VQA tasks are framed as multi-class classification problems, we further incorporated two particularly challenging multi-label VQA tasks pertaining to malocclusion, requiring the model to generate exhaustive diagnoses of facial profiles and intricate malocclusion types (Figure~\ref{fig:figure1}.a, Extended Table 1-5, Supplementary Note 1). 

We constructed a large-scale dataset tailored to these two major categories to train DentVLM. For the oral diseases domain, our dataset comprises 93,500 images, encompassing panoramic X-rays and five intraoral image types. For malocclusion, the dataset includes 90,471 images spanning all seven aforementioned modalities. This extensive dataset was acquired from hospitals and clinics across 31 provinces in China between 2017 and 2024, including a cohort of 20,741 patients. A detailed preprocessing pipeline is described in the Methods section (Figure~\ref{fig:figure1}.b). Leveraging these images and their corresponding diagnostic records, we constructed a large-scale Chinese (ZH) VQA dataset with a predefined list of questions (Extended Table 6), comprising 733,338 pairs for oral disease and 495,704 pairs for malocclusion. For precise localization of tooth-related pathologies, the tooth arch was segmented into six distinct regions. From this expansive dataset, a select subset of 2,865 oral disease images and 3,433 malocclusion images were chosen to generate diagnostic rationales utilizing GPT-4o~\cite{gpt4o} (Extended Table 7). This initiative resulted in the creation of 26,397 VQA pairs with rationales for oral diseases and an additional 17,911 pairs with rationales for malocclusion. Furthermore, to enhance the dataset's linguistic diversity, these Chinese VQA pairs were subsequently translated into English (EN), effectively doubling the dataset's size (Figure~\ref{fig:figure1}.b, Extended Table 8). Our efforts yielded a large-scale dataset of 1.47 million bilingual VQA pairs for oral diseases and 0.99 million pairs for malocclusion.

\subsection{Model overview and experimental setup}
DentVLM is designed to enable comprehensive understanding and reasoning across multimodal oral images (Figure~\ref{fig:figure2}.a). It encodes both visual and textual inputs by leveraging the Qwen2-VL~\cite{qwen2vl} as base model (Extended Figure 1.a). Given an input pair consisting of a visual image $I$ and a textual question $Q$, DentVLM employs a specialized vision encoder to digest images at dynamic resolutions. The resulting patched image embeddings are then concatenated with the token embeddings of the textual questions. This combined representation is subsequently fed into a large language model (LLM) decoder, which generates the desired output answers and their corresponding rationales in an autoregressive manner. 

We introduce a two-stage training strategy to progressively enhance DentVLM's performance (Figure~\ref{fig:figure2}.a). In the first stage, both the vision encoder and the LLM decoder are trained using a dataset of simple VQA triplets, denoted as $D_1={(I, Q, A)}$. This phase focuses on rapidly aligning visual and textual representations for quick decision-making. The second stage involves a more refined finetuning of the LLM decoder. We utilize a meticulously curated, high-quality rational VQA dataset, $D_2=(I, Q, A, R, L)$, which extends beyond simple answers ($A$) to include corresponding reasoning rationales ($R$) and precise anatomical locations ($L$). This phase incorporates instruction tuning, chain-of-thought guidance, and structured location-aware supervision, all designed to enhance the model's clinical reasoning capabilities. Consequently, DentVLM is capable of generating not only accurate answers but also explicit reasoning traces, thereby offering more trustworthy decision support for practical oral healthcare applications, see details in the Methods section. 

For the evaluation of DentVLM and baseline models, we assembled a bilingual test set comprising 910 patients, 2,695 images, and 34,942 VQA pairs, which covers all 36 distinct tasks (Figure~\ref{fig:figure2}.b). Additionally, for the assessment of real-world clinical utility, we constructed a separate retrospective dataset, which includes 1,946 patients, 2,642 images, and 3,105 QA pairs. These data were annotated by four senior dentists, each possessing over 8 years of clinical experience (Extended Figure 1.b-c).  We provide detailed biometric statistics and data distribution for both the two-stage training set and the test set, as well as for the clinical study set (Figure~\ref{fig:figure2}.c, Extended Figure 2, 3.a-i, Supplementary Note 2). Furthermore, we report the perceived difficulty, annotation statistics, and confidence levels of the annotated samples within the clinical study, based on expert dentist assessments (Extended Figure 3.c,f-i). The evaluation metrics are detailed in Methods.

\subsection{Overall performance of DentVLM and baselines}
DentVLM was first comprehensively evaluated on the test set against a diverse array of existing multimodal large language models (MLLMs). The baselines included proprietary general-purpose MLLMs, such as GPT-4o~\cite{gpt4o} and the Gemini~\cite{gemini} series, as well as open-source general-purpose MLLMs, including DeepSeek-VL2~\cite{deepseekvl}, Gemma3-4B~\cite{gemma3}, InterVL-2.5-8B~\cite{internvl2.5}, LlaVA-OneVision-7B~\cite{llava-onevision}, Llama-3.2-11B~\cite{llama3}, Qwen2-VL-7B/72B~\cite{qwen2vl}, and Qwen2.5-VL-7B/72B~\cite{qwen2.5vl}. Additionally, domain-specific medical MLLMs, namely LLaVA-Med-v1.5~\cite{llavamed}, HuatuoGPT-Vision-7B/34B~\cite{huatuogptvision}, and RadFM~\cite{radfm}, were included in the evaluation (Figure~\ref{fig:figure3}.a, Extended Table 9). 

DentVLM demonstrated significant superiority, surpassing all baseline models across 34 bilingual tasks, also indicating the effectiveness of the training process (Extended Figure 4.a-g). The only exceptions were the midline deviation and sagittal relationship classification tasks, where performance was comparable (Figure~\ref{fig:figure3}.a). We calculated the average performance of each model across different tasks in four distinct settings: diagnosis performance for all oral diseases and malocclusion in both English (EN) and Chinese (ZH) languages. DentVLM consistently achieved average performance ranging between 77.28\% and 78.84\% (95\% confidence interval (CI) 73.46-84.21\%, 71.09-83.47\%, 73.53-84.07\%, 71.12-83.43\% for oral diseases (EN), malocclusion (EN), oral diseases (EN) and malocclusion (ZH), respectively), indicating balanced efficacy across various conditions and exhibiting no discernible bias towards either English or Chinese inputs. On average, DentVLM outperformed the best-performing baseline in each setting by 23.76\% ($P<0.001$) (Figure~\ref{fig:figure3}.b-e). Furthermore, DentVLM is capable of inferring answers and rationales for each VQA item in approximately 1.55 seconds when utilizing powerful computing hardware, which extends to 3.21 seconds on consumer-grade GPUs and could be further optimized in real-world deployments that support higher concurrency (Extended Figure 4.g). These substantial performance improvements underscore the inherent challenges in directly applying existing MLLMs to the nuanced domain of oral healthcare, thereby highlighting the significant practical value of our work to intelligent dentistry. 

Furthermore, we evaluated DentVLM's efficacy in disease localization. For this task, the locations of all oral disease-related conditions were mapped to nine specific regional descriptions (Supplementary Note 1). Each model generated textual descriptions of the detected disease locations, which were then mapped to corresponding dental arch areas to compute the Intersection-over-Union (IoU) (Extended Table 10). DentVLM was compared against four strong baselines: GPT-4o~\cite{gpt4o}, Deepseek-VL2~\cite{deepseekvl}, Qwen2-VL-7B~\cite{qwen2vl}, and Qwen2.5-VL-7B~\cite{qwen2.5vl}, across both EN and ZH languages (Figure~\ref{fig:figure3}.f). DentVLM achieved mean IoU scores of 38.44\% (95\% CI 34.45\%-42.43\%) for English and 40.31\% (95\% CI 35.38\%-45.24\%) for Chinese, surpassing the top-performing GPT-4o by 6.85\% ($P<0.05$) in English and 4.8\% ($P<0.05$) in Chinese for this task.

\subsection{Model generalizability and ablation analysis}
We first investigate the generalizability of our DentVLM architecture against different base models, specifically Qwen2-VL-2B~\cite{qwen2vl}, Qwen2-VL-7B~\cite{qwen2vl}, and Gemma3-4B~\cite{gemma3} (Figure~\ref{fig:figure3}.g). All three DentVLM variants outperformed GPT-4o~\cite{gpt4o} by more than 25\% across four distinct test settings. This demonstrates that our proposed approach allows for flexible integration with base models of varying scales and architectures. The inferior performance of DentVLM variant based on Qwen2-VL-2B aligns with established scaling laws in model parameters. Interestingly, DentVLM with Gemma3-4B performed slightly worse on malocclusion-related tasks but surprisingly surpassed the default version by 6.45\% on oral disease-related tasks. This discrepancy could be attributed to the broader knowledge coverage and enhanced model capabilities of the Gemma3-series models, which were released seven months after Qwen2-VL. We anticipate that the performance of DentVLM will continue to improve with ongoing advancements in base models.

We further investigated the data efficiency of DentVLM by evaluating its performance across different scales of training data. We sampled 1\%, 10\%, and 50\% subsets from our training dataset (2.46 million VQA pairs) to retrain the model. As the volume of training data increased, performance improvements were more pronounced in malocclusion-related tasks and within the Chinese (ZH) language context (Figure~\ref{fig:figure3}.h). This phenomenon could be attributed to a potentially inferior capability of the base model in the Chinese context and the specialized nature of malocclusion, both of which are likely to have received less exposure during the initial pre-training phase.

We assessed DentVLM's zero-shot generalization ability on out-of-distribution dental diseases—conditions that are less frequently observed in real-world clinical scenarios and were not included in the training set. We identified four such tasks: bone islands, embedded tooth, cyst, and veneer (Extended Table 11). Despite the complete absence of direct supervision for these conditions, DentVLM maintained stable diagnostic performance across these rare conditions, achieving average accuracies ranging between 52.03\% and 64.44\% (Figure~\ref{fig:figure3}.i). These compelling results underscore DentVLM's potential for application in broader multimodal and multi-task clinical settings.

We conducted extensive ablation studies to evaluate the impact of key hyperparameters and training strategies on DentVLM's overall performance (Supplementary Note 3, Extended Figure 4.a-i). Optimal training epochs were precisely determined based on observed performance plateaus (Extended Figure 4.c). Crucially, full-parameter fine-tuning demonstrably outperformed Low-Rank Adaptation~\cite{lora} (LoRA), unequivocally validating the efficacy of our chosen optimization approach (Extended Figure 4.d). Training the model exclusively on either oral disease data or malocclusion data invariably failed to achieve competitive performance, thereby underscoring the absolute necessity of mixed training across both critical diagnostic categories (Extended Figure 4.e). In contrast, models trained solely in English (EN) or Chinese (ZH) language exhibited smaller performance differentials, although bilingual training consistently yielded the most superior results (Extended Figure 4.f). Finally, DentVLM maintained remarkably consistent performance during inference across varying image resolutions (Extended Figure 4.h) and diverse instruction phrasings (Extended Figure 4.i), confirming its robustness to these important operational variables.

A comprehensive analysis was conducted to evaluate DentVLM's performance across imaging modalities and diagnostic tasks. DentVLM consistently showed comparable diagnostic accuracy in English and Chinese across all modalities (Extended Figure 5.a). Performance on LAT X-rays was marginally superior (average accuracy: 81.17\% EN, 81.83\% ZH), likely due to LAT images often involving assessment of occlusal relationships, maxillomandibular abnormalities, and skeletal classifications. Disease-specific diagnoses across seven modalities revealed robust average accuracy (77.05\% to 81.83\%) per modality, though 1 to 5 challenging tasks per modality had accuracy below 70\% (Extended Figure 5.b-h). We also evaluated DentVLM's efficacy for diseases diagnosable from multiple modalities (Extended Figure 6.a,b). Performance remained consistent across corresponding modalities for most diseases, with minimal differences between highly similar intraoral views (left/right and upper/lower arches), though inconsistent diagnostic performance persisted for specific tasks across modalities, such as low accuracies for calculus, caries, and midline deviation on panoramic X-rays.

\subsection{Clinical study and human evaluation}
To assess DentVLM's real-world clinical applicability, we orchestrated a series of clinical user studies and human evaluations to scrutinize its capabilities across three critical dimensions: diagnostic effectiveness, the efficacy of DentVLM-aided decision-making, and the intrinsic quality of its generated responses  (Figure~\ref{fig:figure4}, Extended Figure 7, 8, 9, 10). Our study recruited 25 dentists, segmenting them into two distinct groups: 13 junior dentists (Dentist-J) with 1-3 years of clinical experience; and 12 senior dentists (Dentist-S), with over 3 years of clinical experience. Robust checks for self-consistency and group-consistency were performed where both junior and senior practitioners demonstrated average self- and group-consistencies over 83\% across all dentists and subgroups (Extended Figure 7, Supplementary Note 4). 

Our initial analysis compared DentVLM's diagnostic accuracy and efficiency against dentists (Figure~\ref{fig:figure4}.a). DentVLM achieved a mean accuracy of 78.26\% (95\% CI 76.16\%-80.36\%) for oral disease-related tasks, even surpassing the average accuracy of junior dentists ($P<0.001$) and competing with senior dentist groups (95\% CI 75.00\%-79.45\%) (Figure~\ref{fig:figure4}.b). In contrast, for malocclusion-related multi-class tasks, which demand more domain knowledge, DentVLM's performance was marginally lower than that of junior dentists (95\% CI 77.19-81.34\% for DentVLM, and 79.61-83.74\% for Dentist-J), and lagged behind senior dentists by 6.18\% ($P<0.001$) (Figure~\ref{fig:figure4}.b). In challenging multi-label malocclusion tasks, DentVLM performed on par with junior dentists but exhibited lower performance than senior practitioners (95\% CI 47.49-62.87\% for DentVLM, 47.36-63.60\% for Dentist-J, and 51.40-67.12\% for Dentist-S) (Figure~\ref{fig:figure4}.b). A granular analysis across all 36 tasks revealed that DentVLM underperformed both levels of dentists in only 4 oral disease-related tasks. Conversely, it outperformed senior dentists in 3 malocclusion tasks and junior dentists in 8 malocclusion tasks (Extended Figure 8.a-b). 

We then investigated the synergistic potential of DentVLM in assisting dentists with diagnoses. Two assistance scenarios were explored: 1) DentVLM's predicted answers were provided to dentists during diagnosis (DentVLM-A); and 2) both DentVLM's predicted answers and rationales were provided (DentVLM-R) (Figure~\ref{fig:figure4}.a). For oral disease diagnosis, both DentVLM-A and DentVLM-R significantly enhanced the performance of junior dentists from 72.25\% to a range of 77.90-79.29\% (all with $P<0.001$). Similarly, senior dentists' performance improved from 77.23\% to 81.43-81.62\% (all with $P< 0.001$), demonstrating consistent effectiveness across both cohorts (Figure~\ref{fig:figure4}.c). In multi-class malocclusion diagnosis, DentVLM-A/R slightly boosted junior dentists' performance by 1.67-3.36\% (95\% CI 79.61-83.75\% for Dentist-J, 81.36-85.33\% for DentVLM-A, 83.14-86.94\% for DentVLM-R) and senior dentists' performance by 1.4-1.98\% (95\% CI 83.57-87.33\% for Dentist-S, 85.66-89.20\% for DentVLM-A, and 85.05-88.66\% for DentVLM-R) (Figure~\ref{fig:figure4}.c). These improvements align with DentVLM's stand-alone performance, where it often surpassed dentists in oral disease tasks but underperformed in certain malocclusion conditions (Extended Figure 9.a-b). For challenging multi-label tasks, while DentVLM-A provided minimal aided performance for junior dentists (95\% CI 47.26-63.60\% for Dentist-J, 47.70-63.90\% for DentVLM-A), DentVLM-R significantly boosted their accuracy by 5.06\% (95\% CI 52.57-68.42\% for DentVLM-R). Conversely, for senior dentists, both DentVLM-A and DentVLM-R yielded substantial improvements, ranging from 7.53-8.64\% (95\% CI 51.40-67.12\% for Dentist-S, 60.37-75.43\% for DentVLM-A, and 59.29-74.30\% for DentVLM-R) (Figure~\ref{fig:figure4}.c). Such pronounced improvements, particularly for senior dentists, may stem from the inherent complexity of multi-label tasks, which demand comprehensive diagnoses that even experts might occasionally overlook. The more evident gains among senior dentists could be attributed to their heightened experience and capacity to optimally leverage AI-driven assistance (Extended Figure 9.e-f). 

DentVLM's inference latency was measured on the powerful GPU with 80G VRAM. Its diagnostic time for multi-class oral disease, multi-class malocclusion, and multi-label malocclusion tasks constituted merely 22.31\% ($P<0.001$), 25.52\% ($P<0.001$), and 15.75\% ($P< 0.05$) of senior dentists, respectively, and remains consistent across tasks (Figure~\ref{fig:figure4}.d, Extended Figure 8.c-d). We also recorded the time expenditure for dentists when assisted by DentVLM-A and DentVLM-R using our web application system for clinical tests (Figure~\ref{fig:figure4}.e, Extended Figure 9.c-d,g-h). Our observations reveal that both DentVLM-A and DentVLM-R profoundly enhanced the efficiency of junior and senior dentists (all with $P<0.001$). 

Finally, we qualitatively evaluate the quality of answers and rationales generated by DentVLM from six distinct perspectives, involving both junior and senior dentists. These included the accuracy of diagnostic results and the correctness, completeness, fairness, faithfulness, and overall acceptability of the generated rationales. Cases that did not yield meaningful diagnostic results were excluded from this analysis. Both junior and senior dentists consistently assigned high ratings across all six dimensions for oral diseases and malocclusion diagnosis, demonstrating strong clinical acceptance of DentVLM's reasoning capability (Figure~\ref{fig:figure4}.f, Extended Table 12, 13).

\subsection{Practical utility of DentVLM in real-world deployment }
\begin{figure}
	\centering
	\includegraphics[width=1\linewidth]{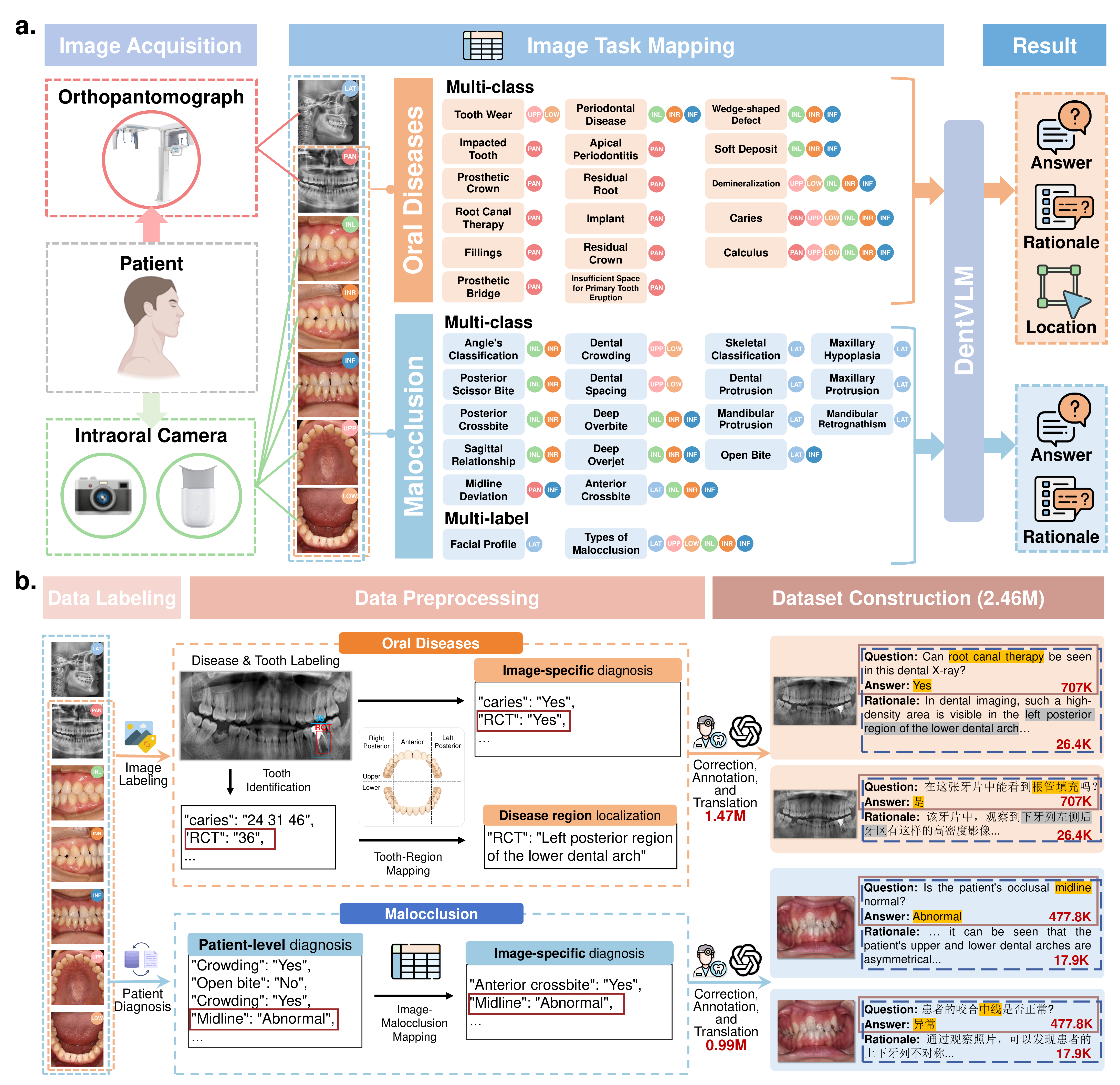}
	\vspace{0.01cm}
	\caption{\textbf{Problem formulation and dataset construction flowchart of the DentVLM.}
		\textbf{a.} The DentVLM diagnostic workflow comprises three stages: image acquisition, image-task mapping, and result generation. Image acquisition involves collecting 7 multimodal oral images (e.g., X-rays, intraoral camera images). Image-task mapping assigns 17 tasks of multi-class oral disease, as well as 2 multi-label and 17 multi-class malocclusion tasks to specific imaging modalities. Finally, DentVLM can generate high-quality, interpretable textual outputs including diagnostic answers, rationale, and location. 
		\textbf{b.} Dataset processing with three phases: data labeling, data preprocessing, and dataset construction. Data labeling yields coarse structured labels via disease and tooth localization for oral diseases related tasks and patient-level diagnosis for malocclusion related tasks. Data preprocessing refines coarse labels into modality-specific diagnoses and disease region localizations, with tooth numbers mapped to six major oral regions. Dataset construction first forms diagnostic VQA pairs without rationale, where a subset undergoes expert fine-grained correction and annotation by GPT-4o to incorporate rationale. GPT-4o-mini is utilized to create a 2.46 million bilingual (EN and ZH) VQA dataset.}
	\label{fig:figure1}
\end{figure}

We investigated the practical utility of DentVLM across three real-world scenarios (Figure~\ref{fig:figure5}.a-f). First, we considered the increasingly prevalent scenario where oral images are captured by edge devices, such as smartphone or portable equipment, which are ideal for population oral health management (Figure~\ref{fig:figure5}.a). We deployed DentVLM to generate diagnosis results across seven oral diseases related to intraoral images (Figure~\ref{fig:figure1}.a, Extended Table 6). A comprehensive list of diagnosed diseases is then aggregated using either majority voting or matching voting (Supplementary Note 5). On a held-out set of 61 patients, DentVLM achieved a diagnosis matching score of 93.91\% for health management using matching voting, representing the upper bound of the model's capability in predicting patient-level disease lists. With majority voting, a diagnosis matching score of 78.22\% was achieved, providing more convincing and reliable diagnostic outcomes (Figure~\ref{fig:figure5}.c). 

Second, we applied DentVLM to simulate real-world orthodontics diagnosis scenarios within a hospital setting (Figure~\ref{fig:figure5}.b). We collected medical images captured using professional equipment across all seven modalities and queried DentVLM to obtain a comprehensive list of diagnosed malocclusion diseases. Results from an external test set involving 69 patients showed that DentVLM achieved a diagnosis matching score of 88.14\% with matching voting and 85.33\% with majority voting (Figure~\ref{fig:figure5}.c). Such performance shows promise for streamlining diagnostic workflows and reducing the rate of missed diagnosis. 

Finally, we explored integrating a reasoning model, DeepSeek-R1~\cite{deepseekr1}, with DentVLM, constructing a preliminary multi-agent collaborative framework to provide more interpretable outputs and user-friendly interactions (Figure~\ref{fig:figure5}.d). We sampled 50 data items from the clinical study set, comprising 50 images covering all modalities and 31 tasks. Each image was initially processed by DentVLM to generate diagnoses and reasoning rationales, which were then prompted to DeepSeek-R1 for enhanced clarity and multiple rounds of interaction (Extended Table 14). Three professional dentists, each with over 8 years of experience, were invited to evaluate these interactions across seven distinct dimensions (Figure~\ref{fig:figure5}.e). All dimensions received scores above 3, with fairness, acceptability, and readability approaching 4, demonstrating the strong practical utility of DentVLM (Extended Figure 11, 12, 13). 

\section{Discussion}
The inherent complexity of oral diseases demands a paradigm shift toward holistic, patient-centered diagnostics and management. Despite advances in AI technology, clinical deployment, particularly in dentistry, remains fragmented and constrained by narrow task-specific models. Overcoming this fragmentation requires developing AI foundational  models capable of unified, multimodal analysis for universal pan-specialty disease evaluation. Such technology promises not only to revolutionize oral healthcare through integrated diagnostic pathways but also to establish a foundational framework for AI-driven clinical reasoning across specialties. By delivering precision diagnostics to resource-constrained settings, this approach offers a practical solution to critical global-health gaps, such as the lack of timely, accurate testing, and equips clinicians in underserved regions with expert-level decision support.

\begin{figure}
	\centering
	\includegraphics[width=1\linewidth]{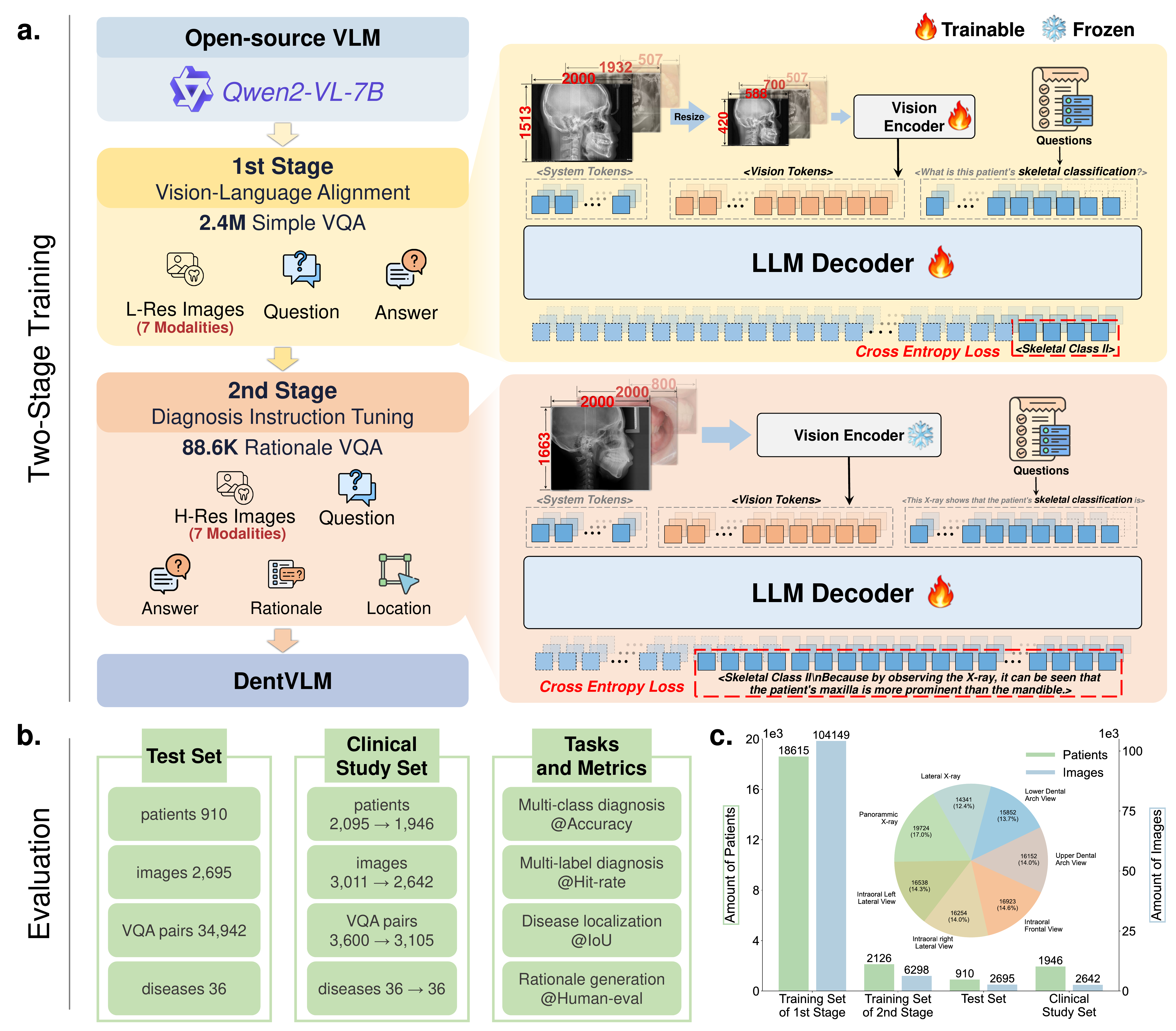}
	\vspace{0.01cm}
	\caption{\textbf{Training and evaluation process of DentVLM.}
		\textbf{a.} DentVLM employs a two-stage training framework built upon the open-source MLLM (i.e., Qwen2-VL-7B). The first stage, Vision-Language Alignment, trains on 2.4 million low-resolution images (spanning 7 modalities) paired with simple VQA pairs to enhance question-answering and visual comprehension. During this phase, both the visual encoder and language decoder are trainable. The second stage, Diagnosis Instruction Tuning, involves further training on 88.6K high-quality, high-resolution VQA data, inclusive of diagnostic results, reasoning processes (rationale), and localization information. In this stage, only the language decoder is fine-tuned while the visual encoder remains frozen, aiming to improve the model's specialized diagnostic and reasoning capabilities.
		\textbf{b.} The bilingual test set encompasses 2,695 images from 910 patients, covering 36 distinct diseases, and totals 34,942 VQA pairs. The clinical study set encompasses 2,642 images from 1,946 patients, covering 36 distinct diseases, and totals 3,105 VQA pairs. Evaluation tasks include multi-class diagnosis (metric: accuracy), multi-label diagnosis (metric: hit-rate), disease localization (metric: Intersection over Union, IoU), and rationale generation (assessed via human evaluation and correction).
		\textbf{c.} Statistics detailing patient (green) and image counts (blue) for each dataset: the first stage training set, the second stage training set, test set, and clinical study set. The distribution of different oral and maxillofacial image types (totaling 115,784 images, including lateral cephalometric, panoramic, and various intraoral views) within all four datasets is presented, with image type percentages depicted in a pie chart.}
	\label{fig:figure2}
\end{figure}

In this work, we developed DentVLM, a unified vision-language model for comprehensive diagnosis of 36 distinct dental conditions across seven imaging modalities. DentVLM was backed by an extensive dataset of 2.46 million dental VQA pairs, including 88,616 with high-quality annotation and rationales. A core technical innovation of DentVLM is its two-stage training strategy, which optimizes the use of large-scale, mixed-quality medical datasets for unified VLM development (Figure~\ref{fig:figure2}.a, Extended Figure 4.a,b). The first stage rapidly aligns the model to the dental domain using 2.37 million simple VQA pairs, with full parameter updates applied to both the vision encoder and language decoder, establishing a broad foundational understanding. The second stage refines the model's capabilities with a smaller (88,616) dataset of high-quality VQA pairs augmented with rich reasoning rationales. This stage significantly enhances diagnostic accuracy and fosters interpretable decision-making, which is indispensable for robust clinical dentistry. Comprehensive ablation studies confirm the necessity of both stages, demonstrating superior performance compared to using either stage in isolation (Extended Figure 4.a). These results underscore the contributions of both large-scale alignment data and high-quality, rationale-enhanced data to overall model efficacy. This strategy represents an efficient approach to maximizing performance while minimizing the costly burden of extensive expert annotation, offering insights for future VLM development in medicine and other domains.

DentVLM addresses a diverse range of diagnostic tasks, overcoming limitations of prior research that focused on narrow applications within specific modalities, such as tooth segmentation in intraoral scans or cone-beam CT~\cite{cui2019toothnet,hao2022toward,xiong2023tsegformer,qiu2022darch,cui2021tsegnet,wu2022two,chung2020pose,jang2021fully,cui2022fully,liu2022hierarchical}, tooth alignment in intraoral scans~\cite{shi2024leta,wei2020tanet,lingchen2020iorthopredictor,li2020malocclusion,wang2022tooth}, landmark localization in lateral cephalometric X-rays~\cite{schwendicke2021deep,zeng2021cascaded,guo2025towards}, or diagnosis of conditions like caries in panoramic X-rays~\cite{zhu2023artificial,chen2024cariesxrays,hamamci2023diffusion,yang2022histopathology}. These earlier methods typically relied on task-tailored architectures and training strategies. In contrast, DentVLM seamlessly integrates and analyzes all seven major 2D dental imaging modalities, providing comprehensive and accurate diagnoses across multiple diseases. This capability marks a pivotal advance toward a general-purpose model for dental analysis and real-world clinical deployment. Furthermore, DentVLM substantially improves upon existing general-domain VLMs (e.g., Qwen~\cite{qwen2vl,qwen2.5vl}, Gemma~\cite{gemma3}, LLaVA~\cite{llava-onevision,llavamed} series) and medical-domain VLMs (e.g., Huatuo~\cite{huatuogptvision}, RadFM~\cite{radfm}) by demonstrating enhanced understanding and utility within dentistry. Consequently, DentVLM bridges a critical gap by harnessing the generalization potential of pretrained VLMs while surpassing the task-specific focus and limited universal diagnostic capabilities of smaller, specialized models.

To demonstrate the clinical applicability of DentVLM, we presented a detailed analysis comparing its diagnostic capabilities with human dentists.  While individual dentists exhibited greater diagnostic variation during self-consistency assessments, their collective group-consistency showed significantly lower variance (Extended Figure 7.d-e). Both self- and group-consistency metrics improved when dentists used DentVLM for assistance (Extended Figure 7.d). At the disease level, DentVLM underperformed junior and senior dentists on only 4 out of 17 and 8 out of 17 oral disease tasks, respectively (Extended Figure 8.a). For malocclusion diagnosis, DentVLM outperformed junior and senior dentists on 8 out of 19 and 3 out of 19 tasks, respectively (Extended Figure 8.b). This suggests particular promise for DentVLM in orthodontic applications, a domain that typically demands extended specialized training.

DentVLM also showed substantial clinical utility when assisting clinicians. Junior and senior dentists experienced performance degradation both in only 3 out of 36 tasks when aided by DentVLM-A or DentVLM-R (Extended Figure 9.a,b). This, coupled with reduced diagnosis times, highlights DentVLM's potential to improve both diagnostic accuracy and efficiency (Extended Figure 9.c,d). However, DentVLM-R, which provides rationales, underperformed DentVLM-A for junior (16 of 36 tasks) and senior (14 of 36 tasks) dentists (Extended Figure 9.e,f). This suggests that while rationales can aid interpretability, they might also introduce cognitive bias or information overload, particularly for experienced clinicians whose extensive knowledge may lessen the perceived value of AI-generated rationales. Time-cost analysis supported this, showing rationale review increased diagnosis duration across most tasks (Extended Figure 9.g,h). Therefore, future implementations should explore adaptive rationale delivery based on task complexity and user expertise to optimize workflow. During the study, dentists also reported diagnostic confidence and case complexity. Both DentVLM and human dentists showed lower accuracy for VQA pairs with lower confidence, typically associated with ambiguous conditions (Extended Figure 10.a,b). DentVLM consistently improved junior dentists' performance across varying complexity levels. For senior dentists, however, its impact on "easy" cases was less pronounced, possibly due to complex AI cues potentially confusing highly experienced practitioners (Extended Figure 10.c,d). 

To demonstrate DentVLM's real-world utility, we validated its performance across three practical scenarios (Figure~\ref{fig:figure5}.a,b,d). The model achieved a diagnosis matching score of up to 93.91\% in home-based dental health management and 88.14\% in intelligent hospital diagnostics (Figure~\ref{fig:figure5}.c), generating accurate lists of potential diseases or malocclusions from images. For patients, the over 90\% diagnosis matching score in household settings signals a highly accessible and dependable tool for early screening, promising to increase early detection of common oral diseases and reduce treatment costs. For dental practitioners, the nearly 90\% diagnosis matching score in hospitals could streamline diagnostic workflows, decrease missed diagnoses, and free up clinical time for more complex cases and patient communication. These applications position DentVLM to bridge the gap between AI and practical dental care, laying a foundation for intelligent dental ecosystems. Concurrently, integrating DeepSeek-R1~\cite{deepseekr1} with DentVLM to form a multi-agent collaborative framework yielded compelling results. Its user interaction records, evaluated by three experienced dentists, scored over 3 (moderately acceptable) and approached 4 (good performance), confirming its efficacy in delivering user-friendly and clinically valuable responses (Figure~\ref{fig:figure5}.e). This establishes a new paradigm for medical AI, moving beyond analysis and diagnosis to become a trusted, interactive partner in doctor-patient communication, augmenting both clinical efficiency and care quality.

Despite promising results, DentVLM has several limitations that highlight areas for future research and development. First, achieving enhanced performance and broader generalizability requires scaling to additional imaging modalities and integrating larger, higher-quality datasets, especially those with longitudinal annotations. The integration of multimodal data, including clinical texts, structured knowledge bases, and 3D imaging modalities (e.g., cone-beam CT, MRI), will be crucial. Such integration would empower expanded capabilities, span early disease detection, automate treatment planning, and advance prognostic analysis, thereby extending DentVLM's clinical utility to complex fields like maxillofacial surgery and oral oncology. Second, further architectural and training innovations are promising. While the current Qwen2-VL~\cite{qwen2vl} implementation demonstrates strong performance, and ablation studies with Gemma3~\cite{gemma3} suggest potential for further gains, adopting cutting-edge foundation models (e.g., Qwen2.5-VL~\cite{qwen2.5vl}) and advanced training strategies, such as reinforcement learning for enhanced reasoning or sophisticated agentic learning paradigms incorporating real-world feedback, could yield performance improvements. Finally, to establish DentVLM's real-world impact and ensure equitable deployment, it is imperative to conduct multi-center clinical trials across diverse populations (varying in age, gender, ethnicity, and comorbidities). Prioritizing deployment in low-resource regions is essential to directly address specialist shortages. Such comprehensive studies will be fundamental in establishing a paradigm for AI-augmented preventive, diagnostic, and therapeutic solutions within global health systems.

In conclusion, DentVLM demonstrates the potential of multimodal, specialty-specific foundation models for diverse dental clinical needs. Through comprehensive pretraining on vast bilingual dental image-text data and rigorous validation, DentVLM consistently exhibits robust performance, showcasing exceptional diagnostic accuracy, enhanced efficiency, and significant clinical utility. Our development, combining systematic data curation, advanced vision-language alignment, instruction tuning, and stringent clinical validation, establishes a framework for building adaptable medical AI systems. These systems can serve various clinical expertise levels and healthcare settings, from specialized clinics to resource-limited regions. These compelling findings indicate promising directions for developing foundation vision-language models in other medical specialties requiring integrated imaging and complex clinical workflows to optimize patient care.

\clearpage

\begin{figure}
	\centering
	\includegraphics[width=.95\linewidth]{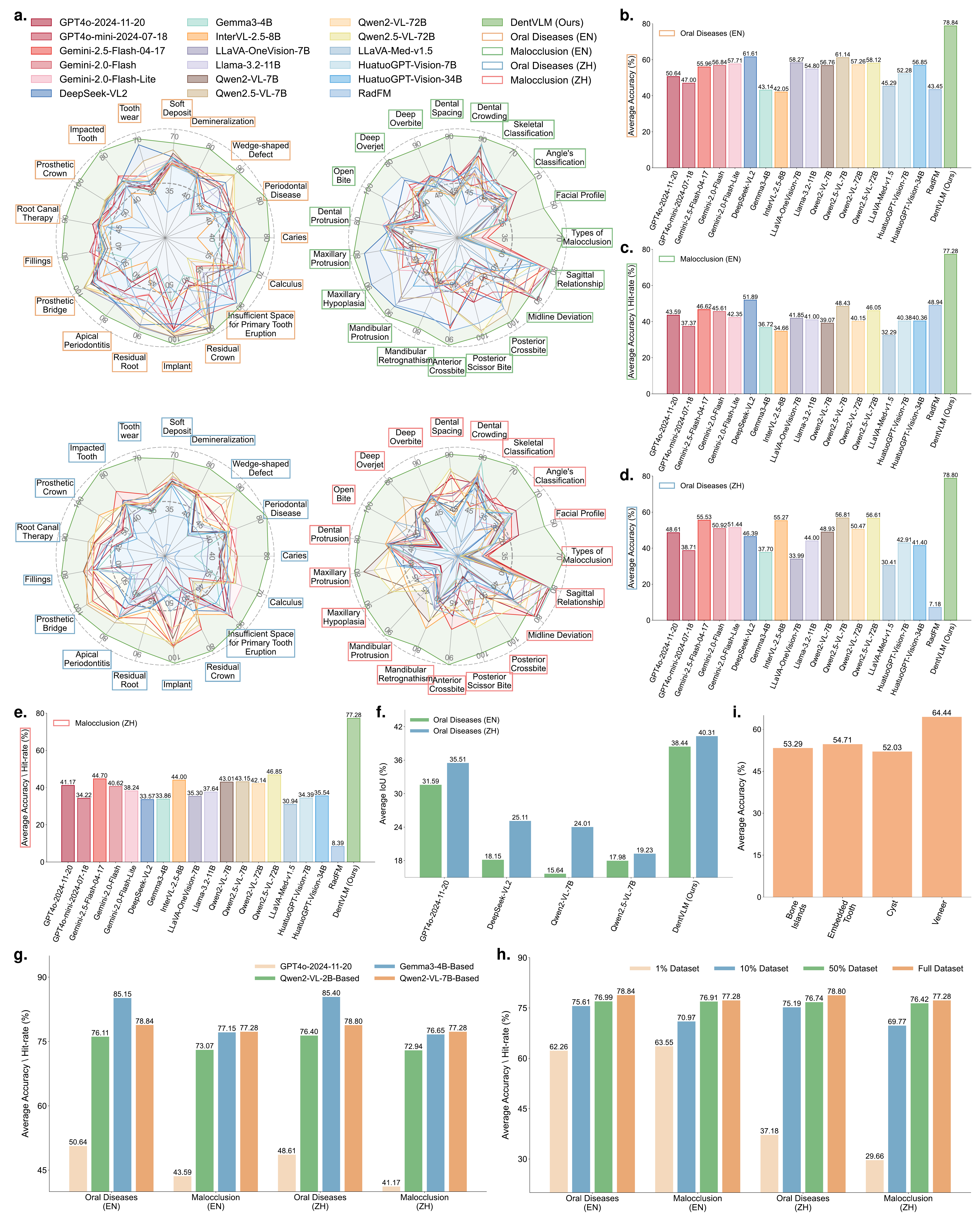}
	\vspace{0.01cm}
	\caption{\textbf{Overall performance evaluation of DentVLM.}
		\textbf{a-e.} Performance comparison of DentVLM and various baselines (proprietary general-purpose MLLMs, open-source general-purpose MLLMs, domain-specific medical MLLMs). Radar charts illustrate model performance on specific oral diseases tasks (left two charts) and malocclusion tasks (right two charts). Bar charts compare the mean accuracy of these models on different categories or language context. All of them demonstrate DentVLM's superior ability, clear advantage and balanced performance comparing baselines.
		\textbf{f.} Comparison of mean IoU scores for disease localization between DentVLM and selected strong baselines on oral disease-related tasks, displaying results for EN (green) and ZH (blue) separately. DentVLM attains the highest scores in both languages (38.44\% EN, 40.31\% ZH), indicating superior  and stable performance across bilingual inputs for disease localization.
		\textbf{g.} Generalizability over base models. Different scale (Qwen2-VL-2B) and architecture (Gemma3-4B) models are conducted on the proposed two-stage training framework.
		\textbf{h.} Generalizability over data scales (1\%, 10\%, 50\% and full dataset).
		\textbf{i.} Generalizability over out-of-distribution diseases, including bone islands, embedded tooth, cyst, and veneer.}
	\label{fig:figure3}
\end{figure}

\begin{figure}
	\centering
	\includegraphics[width=.975\linewidth]{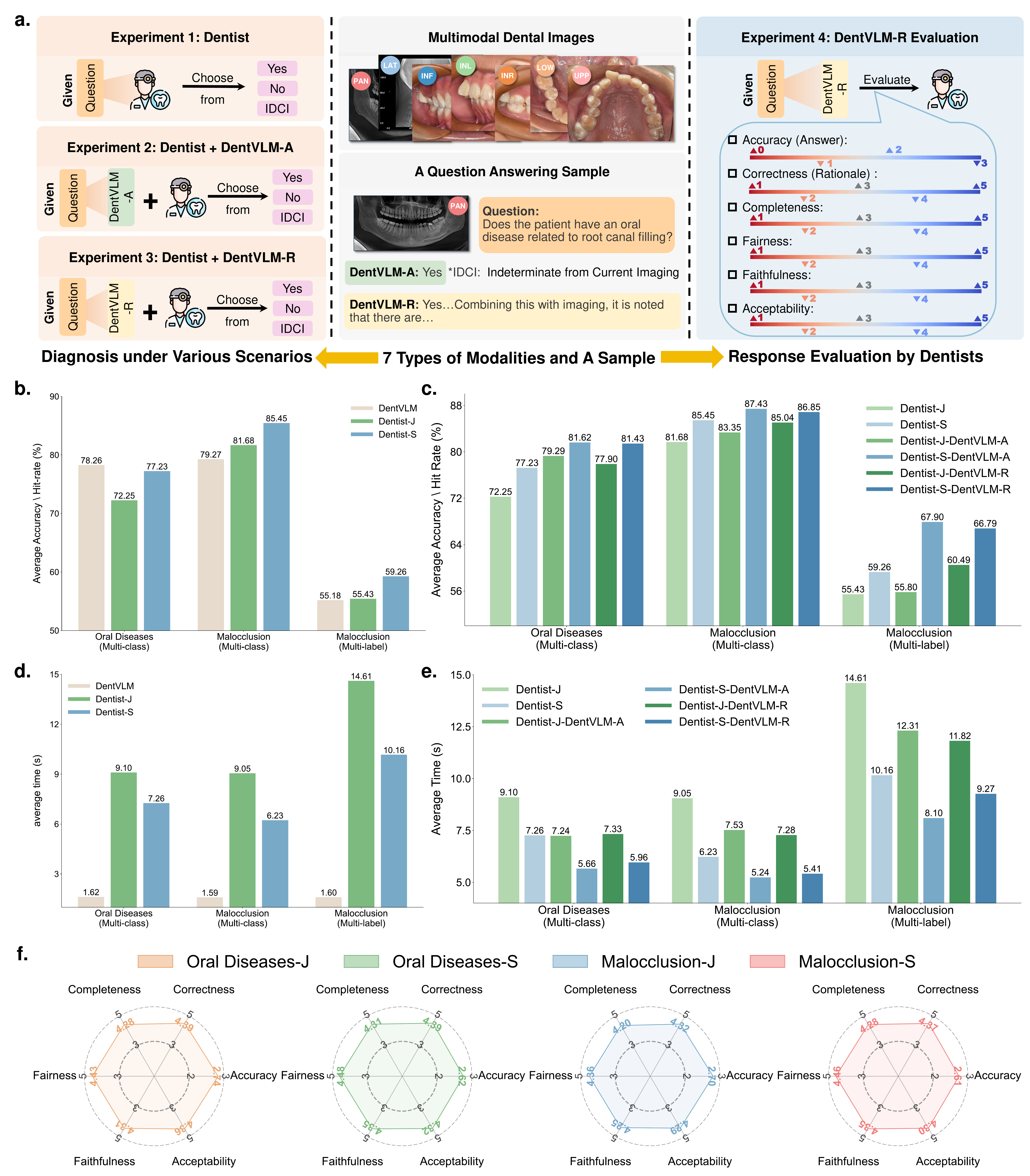}
	\vspace{0.01cm}
	\caption{\textbf{Clinical study of DentVLM.}
		\textbf{a.} Overview of data modalities and experimental content in the clinical study, which include diagnosis under various scenarios and response evaluation by dentists. In experiment 1, dentists performed independent diagnoses based on clinical questions without model assistance. Experiment 2 introduced DentVLM-generated answers (DentVLM-A), which dentists used to inform their judgments. Experiment 3 further provided DentVLM's rationale (DentVLM-R), enabling dentists to make diagnostic choices based on more comprehensive responses from the model. In Experiment 4, dentists subjectively assessed the quality of DentVLM-R's responses across six dimensions: answer accuracy, as well as rationale's correctness, completeness, fairness, faithfulness, and acceptability.
		\textbf{b.} Comparison between DentVLM and dentists of varying expertise in terms of accuracy on multi-class tasks and hit-rate on multi-label tasks.
		\textbf{c.} Comparison of dentists performance under DentVLM-assisted and unassisted conditions.
		\textbf{d.} Diagnosis efficiency of dentists against DentVLM.
		\textbf{e.} Diagnosis efficiency of dentists with different assistance by DentVLM.
		\textbf{f.} Subjective evaluation of DentVLM-generated response across six dimensions by dentists.
		In \textbf{c}, \textbf{e}, ``DentVLM-A'' represents assistance by DentVLM's responses containing only answers, and ``DentVLM-R'' represents assistance by DentVLM's responses with answers and rationales. In \textbf{b}-\textbf{f}, ``-J'' and ``-S'' denote junior and senior dentists, respectively.}
	\label{fig:figure4}
\end{figure}

\begin{figure}
	\centering
	\includegraphics[width=1\linewidth]{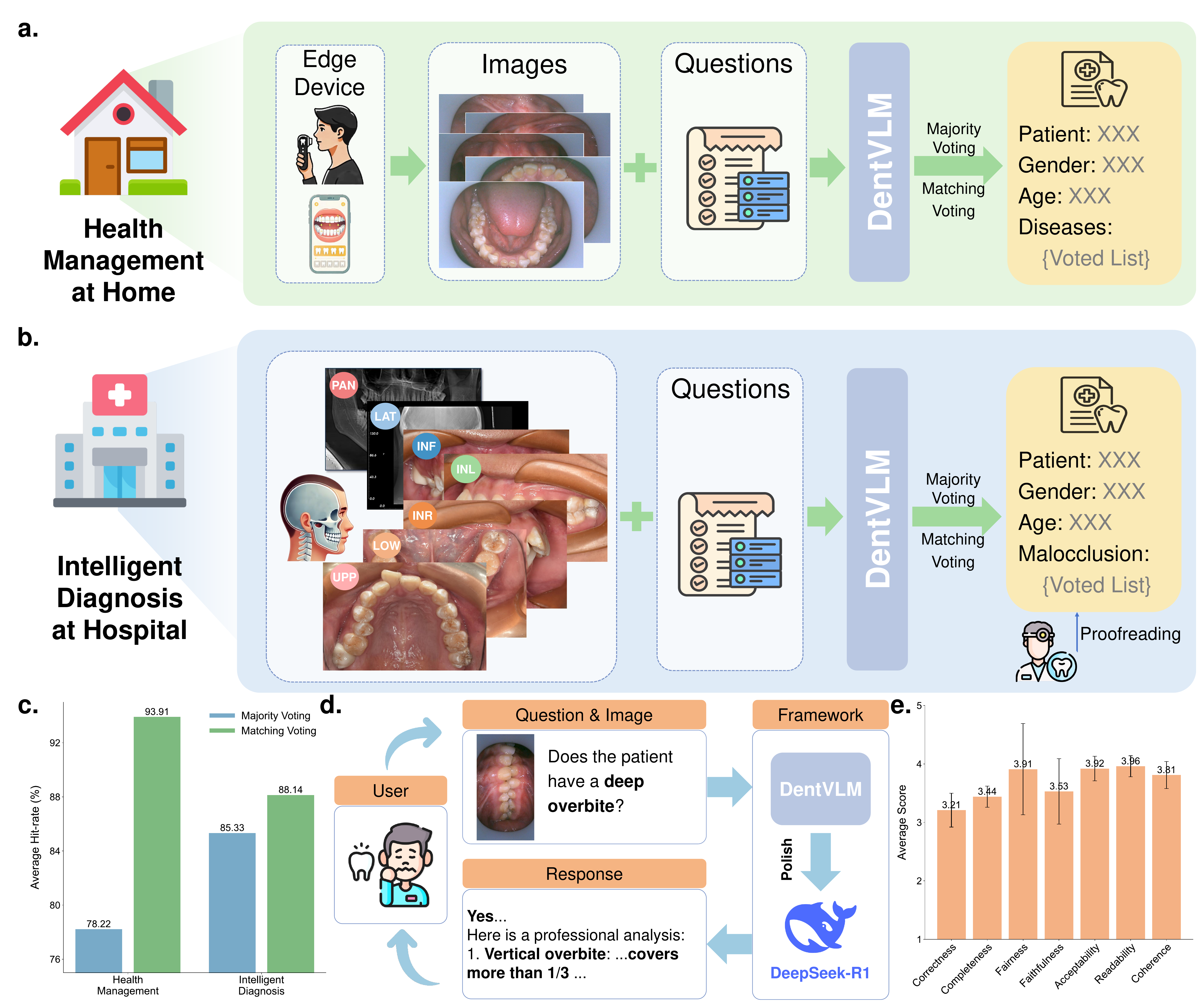}
	\vspace{0.01cm}
	\caption{\textbf{Exploration of practical utility for DentVLM in real-world deployment.}
		\textbf{a.} Schematic diagram of the DentVLM application flow in health management at home. Users capture oral images at home via edge devices (e.g., smartphones, portable oral imaging devices). These images, along with predefined questions, serve as input for DentVLM. Results are aggregated using either majority voting or matching voting to determine a list of potential oral diseases.
		\textbf{b.} Intelligent diagnosis workflow simulation of using DentVLM at hospital. Professional oral images (including panoramic X-rays, cephalometric radiographs, and multi-angle intraoral photographs), combined with predefined questions, are input into the DentVLM. The model analyzes the patient's malocclusion condition, employing majority voting or matching voting, to automatically generate structured diagnostic results, which further undergo clinical proofreading, establishing an intelligent, efficient, and clinically feasible assisted diagnosis workflow.
		\textbf{c.} Performance of different voting strategies for health management at home and intelligent diagnosis at hospital.
		\textbf{d.} Demonstration of the multi-agent collaborative framework integrating DeepSeek-R1, where initial response of DentVLM based on user's question and image would be polished by reasoning model into more interpretable and user-friendly interactions.
		\textbf{e.} Dentists' subjective evaluation for interaction records of multi-agent collaborative framework across seven dimensions.
		In \textbf{a}, \textbf{b}, ``Voted List'' indicates final prediction containing various potential symptoms.}
	\label{fig:figure5}
\end{figure}

\clearpage
\section{Methods}
\subsection{Ethical approval}
This study was approved by the institutional review boards of Sichuan University (approval number: WCHSIRB-D-2021-331) and Shanghai Jiaotong University (approval number: SH9H-2020-TK400-1). The clinical study followed the Declaration of Helsinki protocols and was approved by the research ethics committee of West China Hospital of Stomatology and Shanghai Ninth People's Hospital. All data used for model training, validation, and testing were fully anonymized and authorized by all involved patients. 

\subsection{Training dataset construction}
Based on the formulated problem (Figure~\ref{fig:figure1}.a, Supplementary Note 1), we curated a large-scale dataset comprising 20,741 patients, 110,447 images, and 2,458,084 bilingual VQA pairs to train DentVLM (Figure~\ref{fig:figure1}.b). The patients were sourced from hospitals and dental clinics across 31 provinces in China between 2017 and 2024. The images cover seven modalities: LAT (13,628, 12.3\%), PAN (18,237, 16.5\%), INL (15,818, 14.3\%), INR (15,539, 14.1\%), INF (16,234, 14.7\%), UPP (15,638, 14.2\%), and LOW (15,353, 13.9\%). Collectively, this diverse dataset supports 17 tasks related to oral disease (Extended Table 2) and 19 tasks related to malocclusion (Extended Table 3,4).

\textbf{Preprocessing and construction for tasks of oral diseases.} For tasks related to oral diseases, we collected 93,500 images including PAN and five types of intraoral image, where each annotated with bounding boxes for every disease and tooth regions, as well as FDI notations. Using these spatial annotations, we directly determined the image-specific diagnosis: if a bounding box for a given disease was present, the task was mapped ``yes'' otherwise ``no''. Moreover, we calculated the distance between the bounding boxes and tooth, and constructed the disease region localization based on the tooth-region mapping (Extended Table 5). In particular, the teeth were split into upper and lower parts in the vertical direction based on their respective jaw planes, and into right posterior, anterior, and left posterior in the horizontal direction based on their distance from the midline.  Subsequently, based on the image-specific diagnosis, we randomly sampled questions for every task to construct 733,338 VQA pairs $(I, Q, A)$ in Chinese (ZH) (Extended Table 6). To ensure the quality of the source data used for guiding the rationale generation, we invited 6 expert dentists to correct image-specific diagnosis and disease region localization from 2,865 images. Then, we instructed GPT-4o~\cite{gpt4o} (i.e. gpt-4o-2024-08-06) to annotate reasoning steps based on the provided images, questions, answers, and potential existing locations with predefined prompts for these rechecked images (Extended Table 7). We manually reviewed 5\% of the rationales to ensure fluency and logic. Finally, we obtained 26,397 VQA pairs $(I, Q, A, R, L)$ with additional rationales for 2,865 images and 706,941 VQA pairs $(I, Q, A)$ for the remaining 90,635 images. 

We translate the original ZH dataset to English (EN). For data items without rationale, since the questions originated from a predefined list and the answers fell within a fixed range, we employed an end-to-end dictionary-based approach that covered all possible phrasings for direct matching and translation (Extended Table 2,5,6). In contrast, for VQA pairs with rationale, we used the GPT-4o-mini~\cite{gpt4o} (i.e. gpt-4o-mini-2024-07-18) to guide the rationale translation with elaborate prompts (Extended Table 8). We concatenated the translated answer with the rationale to construct the corresponding EN response. Ultimately, we  generated 1.47 million bilingual VQA pairs for oral disease.

\textbf{Preprocessing and construction for tasks of malocclusion.} For tasks related to malocclusion, we collected 90,471 images from 14,420 patients, encompassing all the 7 modalites and diagnosis reports for each individual, which includes specific manifestations related to malocclusion tasks. We first established modality-specific diagnosis results based on the image-task mapping which was predefined by three senior dentists (Figure~\ref{fig:figure1}.a). Next, we randomly sampled questions to construct 495,704 VQA pairs $(I, Q, A)$ in ZH (Extended Table 6). However, considering the potential misalignment of information when directly mapping the malocclusion issues from the diagnostic report to specific images (e.g. the dental crowding of a patient may occur only in the upper arch while the lower arch remains healthy), we invited 28 orthodontic technicians to correct 17,911 VQA pairs from 557 patients' 3,433 images. The rationales were annotated for these pairs by GPT-4o (i.e. gpt-4o-2024-08-06) based on the provided images, questions and answers with predefined prompts (Extended Table 7). Finally, we obtained 17,911 VQA pairs $(I, Q, A, R)$ with additional rationales for corrected 3,433 images, and 477,793 VQA pairs $(I, Q, A)$ for the remaining 87,038 images. We translated them into EN with the similar procedure described in oral diseases tasks and generated 0.99 million pairs.

\subsection{Details of DentVLM architecture}
We adopted the general-purpose Qwen2-VL-7B~\cite{qwen2vl} as the base model and fine-tuned it on oral healthcare applications (Figure~\ref{fig:figure2}.a). Unlike the conventional architecture of MLLM~\cite{llava} which comprises a vision encoder, a connector, and an LLM decoder, we removed the connector in DentVLM. Instead, it employs multimodal rotational position embedding to unify the encoding of both visual and textual inputs (Extended Figure 1.a). Specifically, DentVLM's vision encoder is a Vision Transformer~\cite{vit} (ViT) with 675M parameters, a depth of 32 layers, a patch size of 14, and a hidden size of 3,584. It supports dynamic resolution encoding~\cite{dehghani2023patch} of images to minimize feature degradation caused by image resizing. Following the vision blocks, a patch merger, implemented as a simple MLP layer, compresses each group of adjacent 2$\times$2 tokens into a single token, thus reducing resource overhead. The LLM decoder is a 7B-parameter language model from the Qwen2~\cite{qwen2} families, with a depth of 28 layers and a hidden size of 3,584. It supports a maximum context length of 32,768 tokens and can process concatenated token sequences of image and text, generating output text in an autoregressive manner.

\subsection{Two-stage training pipeline of DentVLM}
We introduced a two-stage training strategy to progressively enhance model performance on downstream tasks, which is anticipated to fully exploit previously constructed datasets with mix quality and huge volume, and tailor the base model to the oral healthcare domain (Figure~\ref{fig:figure1}.b, Figure~\ref{fig:figure2}.a). In the following sections, we elaborate on this strategy.

\textbf{Stage 1: Vision-language alignment.} Recognizing that oral healthcare is a highly specialized domain posing significant challenges for the direct application of MLLMs, our primary objective is to align visual and textual representations for dental tasks. To achieve this, we integrated the constructed bilingual dataset, covering all 36 tasks and containing only the answers, leading to approximately 2.4 million image-text pairs $(I, Q, A)$, which we refer to as ``simple VQA''. This subset addresses the scarcity of dental images and domain-specific knowledge in the pretraining corpora of general MLLMs. Moreover, guiding the model to only generate answers reduced the learning difficulty. All images were resized to a resolution with the upper bound of 512$\times$512 while maintaining their aspect ratios before being fed into the model to optimize training efficiency. The goal of this stage is to minimize the negative log-likelihood of the predicted results $A'$ based on carefully curated low-resolution simple VQA subset $D_1= {(I, Q, A)}$:

\begin{equation}
	L(\theta; D_1)=-\frac{1}{N}\sum_{i=1}^N\log{\pi_{\theta}(A'_i|(I_i,Q_i))},
\end{equation}

where $\pi$ indicates our model and $\theta$ refers to its parameters. In practice, we implemented full-parameter training across all modules and employed DeepSpeed ZeRO-2~\cite{deepspeed} to partition optimizer states and gradients, effectively reducing GPU memory usage. Training was conducted with the batch size per device as 12, the gradient accumulation step as 16, epochs as 3, learning rate as $2\times10^{-5}$. We employed the cosine annealing and the warm-up strategy with the ratio of 0.1. 

\textbf{Stage 2: Diagnosis instruction tuning.} Following vision–language alignment, the model already possessed preliminary diagnostic competence. However, the inherently black-box nature of its decision-making process remained inadequate for clinical scenarios that demand high interpretability. Consequently, in the second stage for diagnosis instruction tuning, we employed the dataset that includes explicit rationale and potential existed location annotations, producing a rationale VQA subset of roughly 88.6k samples, formatted as $(I, Q, A, R, L)$. This subset guides the model not only to provide an answer $A$ but also to generate a detailed explanation $R$ with corresponding location information $L$. 

Unlike the first stage, which required large-scale data to achieve fast alignment and domain adaptation, the second stage focuses on curating high-quality and credible data to strengthen the model's diagnostic instruction following and reasoning capability. All source images and annotations used in this stage were annotated and double validated by dentists, and we encoded every image at its original resolution to mirror real clinical practice. Furthermore, considering the vision encoder already extracts image features effectively after the previous learning stage, we froze it and only updated the parameters of the LLM decoder, denoted as $\theta'$. The optimization objective is a cross entropy loss for dataset $D_2={(I, Q, A, R, L)}$:

\begin{equation}
	L(\theta'; D_2)=-\frac{1}{N}\sum_{i=1}^N\log{\pi_{\theta}(A'_i,R'_i,L'_i|(I_i,Q_i))},
\end{equation}

In implementation, we performed full-parameter training of the LLM decoder and employed DeepSpeed ZeRO-2~\cite{deepspeed} to reduce GPU memory consumption. Training was conducted with a per-device batch size of 2, gradient accumulation steps of 32, 5 epochs, and an initial learning rate of $1\times10^{-5}$ using the same decay and warm-up schedule as in the first stage.

\subsection{Downstream evaluation details}
To evaluate the capabilities of DentVLM, we curated a bilingual test set consisting of 910 patients, 2,695 images, and 34,942 VQA pairs, covering all 36 tasks of oral diseases and malocclusion (Figure~\ref{fig:figure2}.b). In parallel, we collected a clinical study set comprising 2,095 patients, 3,011 images, and 3,600 VQA pairs to assess DentVLM's practical applicability in real-world scenarios (Figure~\ref{fig:figure2}.b, Extended Figure 1.b). After annotation and filtering by a team of four expert-level dentists (8 years experience), the finalized clinical study set includes 1,946 patients, 2,642 images, 3,105 QA pairs, and maintains coverage of all 36 cases (Extended Figure 1.c). Based on the data definition and formats of labels for various tasks (Extended Table 2-5), we categorized them into four types: multi-class diagnosis, multi-label diagnosis, disease localization, and rationale generation, with specifically designed metrics for each type. In the following section, we provide a detailed explanation of dataset construction, evaluation metrics and experiments details.

\textbf{Test set construction.} For oral diseases related tasks, we collected 1,220 images along with dentist-annotated locations, which includes 532 panoramic radiographs and 688 intraoral photographs. Based on its data acquisition process, each panoramic corresponds to a unique patient, while intraoral images typically cover five modalities per patient. For malocclusion related tasks, we gathered another 1,475 images from 241 patients, each containing all seven modalities along with corresponding malocclusion diagnosis reports. Then, using the same data preprocessing and correction pipeline of training set $D_2$ (Figure~\ref{fig:figure1}.b), we derived the diagnosis and region localization for each image. Finally, according to the predefined question lists (Extended Table 6), we constructed bilingual VQA pairs $(I, Q, A)$ without rationales, including 19,600 pairs for oral diseases and 15,342 pairs for the malocclusion.

\textbf{Clinical study set construction.} In order to construct the clinical study set, we recruited a total of 2,095 patients to obtain medical images through various imaging devices. We sampled 3,011 images to create 3,600 VQA pairs covering all 36 tasks, ensuring that reliable conclusions could be drawn for each task. To ensure the quality of results and reduce the influence of subjective or accidental factors from individual dentists, we divided the entire dataset into two parts (Extended Figure 1.b). The first part $D_{idp}$, contains 92 data items per task, totaling 3,312 VQA pairs, which were evenly assigned to each dentist. The second part $D_{gv}$ consists of four subsets, each containing 72 data items labeled as $D_{gv\{i\}}(i=1,2,3,4)$, which were used to validate the consistency of clinical results in different dentists' groups (Supplementary Note 4). 

As for the ground truth labeling, we invited four expert dentists to annotate each VQA pair (Extended Figure 1.c). Specifically, each dentist was required to provide a diagnostic answer, its confidence and a complexity rating for this case based on the current image and task. Then, we discarded annotations with low confidence and conducted a vote on the remaining results based on the answer. In cases of a tie or if the selected answer indicated ``The answer is indeterminate from current imaging'', the data item was directly removed. Finally, we combined the annotations corresponding to the final answer, selecting the highest confidence and the most difficult complexity as supplementary annotations. Through this meticulously designed annotation and filtering process, we retained 3,105 VQA pairs $(I, Q, A)$ from 2,642 images of 1,946 patients.

\textbf{Evaluation tasks and metrics.} We categorized the 36 tasks into four groups with different evaluation metrics (Figure~\ref{fig:figure2}.b). \textbf{1) Multi-class diagnosis} includes 34 tasks (Extended Table 2, 3), excluding facial profile and types of malocclusion. These tasks involve determining the presence or specific category of the particular disease, leading to clear and unique answers. Therefore, we use accuracy as the evaluation metric. It is calculated based on the ground truth $A$ and the generated answer $A'$: $\text{Accuracy}_t=\frac{\sum_{i=1}^N{\textbf{1}_{A'_i=A_i}}}{N}$, where $t$ denotes the specific task, and $N$ represents the number of samples for task $t$ in the test set. \textbf{2) Multi-label diagnosis} primarily focuses on facial profile and types of malocclusion (Extended Table 4), where the answers may include multiple conditions. And the model is required to generate an answer list that matches the labels. Thus, we propose the hit-rate, calculated based on the ground truth diseases list $A$ and the $A'$ generated by DentVLM: $\text{Hit-rate}_t=\frac{1}{N}\sum_{i=1}^N\frac{n_2}{n_1}$, where $n_1$ and $n_2$ represent the number of diseases existed in $A$ and $A'$ respectively, and $N$ is the number of samples for task $t$. We set hit-rate to 0 if $A'_i\subseteq A_i$ is not satisfied, i.e., we did not allow any misdiagnosis. \textbf{3) Disease localization} includes 17 tasks related to oral diseases, each specifying one or more regions where the disease exists (Extended Table 2, 5). The model is expected to identify these locations where lesions may exist. For evaluation, we adopt the Intersection over Union (IoU) metric, calculated based on the ground truth location $L$ and DentVLM's generation $L'$: $\text{IoU}_t=\frac{1}{N}\sum_{i=1}^N\frac{\sum_{j=1}^{9}\textbf{1}_{(l_j\in L)\&(l_j\in L')}}{\sum_{j=1}^{9}\textbf{1}_{(l_j\in L)|(l_j\in L')}}$, where $l_j$ is one of the predefined descriptions of location information (Supplementary Note 1). \textbf{4) Rationale generation} encompasses all 36 tasks (Extended Table 2-4). Unlike the previous categories with definite answers, this mainly concentrated on the scenario of open-ended QA, which we assess by human evaluation score. Considering the expertise required for clinical diagnosis, we invite dentists to score responses generated by the DentVLM from multiple dimensions in the clinical study.

\textbf{Statistical analysis.} In overall performance comparison of DentVLM and baselines, as well as across various diagnostic scenarios in the clinical study, we conducted t-tests to determine whether the differences observed were statistically significant. The null hypothesis posits no discernible difference between two competitors and $P<0.05$ was regarded as statistically significant. Additionally, we calculated the 95\% confidence interval (CI) to assess the uncertainty around the mean accuracy or hit-rate. We first computed the sample mean, followed by the standard error of the mean, which was derived from the sample's standard deviation divided by the square root of the sample size. The margin of error was then determined by multiplying the standard error by the critical value from the Student's t-distribution corresponding to a 95\% confidence (i.e. 1.96). This margin was finally added to and subtracted from the sample mean to establish the upper and lower bounds of 95\% CI.

\textbf{Overall performance comparison.} First, DentVLM was evaluated on the test set by comparing against commonly used proprietary general-purpose MLLMs (i.e. GPT-4o-2024-11-20~\cite{gpt4o}, GPT-4o-mini-2024-07-18~\cite{gpt4o}, Gemini-2.5-Flash-04-17~\cite{gemini}, Gemini-2.0-Flash~\cite{gemini}, and Gemini-2.0-Flash-Lite~\cite{gemini}), open-source MLLMs (i.e. DeepSeek-VL2~\cite{deepseekvl}, Gemma3-4B~\cite{gemma3}, InterVL-2.5-8B~\cite{internvl2.5}, LLaVA-OneVision-7B~\cite{llava-onevision}, Llama-3.2-11B~\cite{llama3}, Qwen2-VL-7B/72B~\cite{qwen2vl}, and Qwen2.5-VL-7B/72B~\cite{qwen2.5vl}), as well as domain specific medical MLLMs (i.e. LLaVA-Med-v1.5~\cite{llavamed}, HuatuoGPT-Vision-7B/34B~\cite{huatuogptvision}, and RadFM~\cite{radfm}) (Figure~\ref{fig:figure3}.a-e). For the proprietary models, testing was conducted through their respective official API services. The open-source models were evaluated locally by their released weights. We adopted a two-step inference strategy for these baselines. The original image and question were first input into the model to obtain an initial response. Subsequently, a task-specific prompt was used to guide the model to extract the final answer from the response for evaluation. The prompt was customized based on the language of the question and the expected format of the task's answer (Extended Table 9). For DentVLM, we directly input the image and the corresponding question to prompt the model to generate the final answer along with its rationale. We utilized the vLLM~\cite{vllm} library to accelerate inference, with the batch size as 1. The minimum and maximum pixel values were set to 4$\times$28$\times$28 and 8192$\times$28$\times$28 respectively. The model was deployed independently on a single GPU, supporting a maximum input token length of 16,384 and a maximum output length of 512 tokens. The temperature was set to 0.1.

\textbf{Oral disease location.} We compared DentVLM with GPT-4o-2024-11-20~\cite{gpt4o}, DeepSeek-VL2~\cite{deepseekvl}, Qwen2-VL-7B~\cite{qwen2vl}, and Qwen2.5-VL-7B~\cite{qwen2.5vl} for disease localization. Following the same two-stage inference strategy as in previous experiments, we guided the baseline models to output combinations of the nine predefined location descriptors to compute the final IoU scores (Supplementary Note 1). In detail, we first used the original image along with the bilingual prompts to query the MLLMs. Next, based on the initial response, we further guided the MLLMs to extract the final answer using the corresponding instruction (Extended Table 10). For DentVLM, these descriptors were directly extracted from the generated rationale using an exact match strategy. The IoUs are computed on the 5,749 cases in which DentVLM correctly diagnosed the disease and the condition was indeed present. 

\textbf{Inference across different hardware.} We also examined DentVLM's inference latency across different GPU hardware using the clinical study set (Extended Figure 4.g). We deployed the DentVLM on a single GPU and set the batch size as 1 to closely mimic a real diagnostic process, where questions were answered sequentially, although larger batch sizes can significantly reduce total processing time. Furthermore, we optimized this process using the vLLM~\cite{vllm} library and employed the same parameter configurations (i.e. threshold of pixel values, input token length, output length, and temperature) for all evaluations.

\textbf{Generalization over base models.} We evaluate the adaptability of our training pipeline with additional base models, such as Qwen2-VL-2B~\cite{qwen2vl} and Gemma3-4B~\cite{gemma3} (Figure~\ref{fig:figure3}.g). Compared with the original base model Qwen2-VL-7B, Qwen2-VL-2B has a similar architecture but different scale, where the hidden sizes are both 1536 in vision encoder and LLM decoder. Gemma3-4B follows the general three-component architecture of MLLM, which includes a vision encoder~\cite{sigmoid}, a connector and an LLM decoder. The vision encoder is a ViT to extract visual features with approximately 417M parameters, a depth of 27 layers, a patch size of 14, a hidden size of 1,152 and a fixed input image resolution of 896$\times$896. The connector is a simple projection layer that maps visual features into the latent space of the language model. The LLM decoder has a depth of 34 layers and a hidden size of 2,560, which supports a maximum context length of 128K tokens. We trained all these three models under the same hardware and software environments, using the same training data and hyperparameters configuration, and then performed inference on the test set with the same settings.

\textbf{Generalization over data scales.} We investigate the impact of data scale on DentVLM. From the full training set, we uniformly sampled subsets comprising 1\%, 10\%, and 50\% of the data. Specifically, the 1\% subset contains 23,694 VQA pairs from $D_1$ and 886 VQA pairs from $D_2$; the 10\% subset includes 236,946 and 8,861 VQA pairs from $D_1$ and $D_2$ respectively; and the 50\% subset consists of 1,184,734 and 44,308 VQA pairs from $D_1$ and $D_2$. These subsets were then used to perform two-stage training under the same settings. 

\textbf{Generalization over out-of-distribution diseases.} To evaluate the zero-shot generalization ability of DentVLM, we mined our collected data to identify four relatively rare tasks that were not present in our training set: bone islands, embedded tooth, cyst, and veneer (Extended Table 11). With the same dataset construction pipeline (Figure~\ref{fig:figure1}.b), we obtained 774, 366, 328, and 74 instances for these four tasks in the test set, respectively. We adopted a zero-shot inference approach, where DentVLM generates the final responses only with images and questions.

\textbf{Clinical study setup.} We invited 13 junior and 12 senior dentists to participate in the clinical study (Figure~\ref{fig:figure4}.a), which explored the potential benefits DentVLM brings to dentists of different experience levels. In order to support efficiency and security of these experiments, we also developed a web-based system designed to assist dentists in viewing images, making diagnoses, and giving feedback for model's responses across various devices. The clinical study focused on two primary aspects: a comparative analysis of diagnosis under various settings, and a subjective evaluation of responses generated by our model. For the first aspect, dentists would perform diagnoses under three distinct scenarios: 1) relying solely on their own expertise; 2) assisted by the ``answer-only'' responses generated from the DentVLM-A, 3) assisted by DentVLM-R's responses that included both answers and rationales. When reporting the time cost of this assistance setting, we exclude the inference time of DentVLM. 

Each dentist would evaluate the response from six dimensions: the accuracy of the answer, as well as the rationale's correctness, completeness, fairness, faithfulness, and acceptability. The score of accuracy ranged from 0 to 3, where 0 indicates the question could not be answered based on the current image, 1 indicates completely incorrect, 2 indicates partially correct and 3 indicates totally correct. The scores for the remaining dimensions all range from 1 to 5, with 1 for poor, 2 for fair, 3 for moderate, 4 for good and 5 for excellent. And the last five dimensions respectively emphasize whether the rationale: 1) aligns with diagnostic logic and guidelines; 2) integrates multiple perspectives and information without being one-sided; 3) contains significant biases or safety concerns; 4) avoids fabrications, misleading the user, or nonsensical statements; and 5) the overall acceptance level of the reasoning content.

\subsection{Practical utility for DentVLM in real-world demonstrations}
We explored the potential real-world applications of DentVLM, such as health management at home (Figure ~\ref{fig:figure5}.a), intelligent diagnosis at hospital (Figure ~\ref{fig:figure5}.b) and multi-agent collaborative framework (Figure ~\ref{fig:figure5}.d).

\textbf{Health management at home.} Users capture five intraoral images at home using edge devices (e.g., smartphones, portable oral imaging devices). Then, for each intraoral image and its corresponding oral diseases, we construct VQA pairs based on the image task mapping and predefined question templates (Figure~\ref{fig:figure1}.a, Extended Table 6). Subsequently, DentVLM's responses for each image are consolidated to generate a list that contains all image-specific disease diagnosis results. These lists are aggregated using either majority voting or matching voting to determine the potential diseases list for the current patient, thereby supporting personalized remote preliminary screening (Supplementary Note 5). 

\textbf{Intelligent diagnosis at hospital.} Patients are captured with seven clinical oral images at hospital using professional equipment. For each modality and its corresponding malocclusion tasks, DentVLM generates diagnostic results based on constructed VQA pairs similar to the health management scenario. The diagnosis lists are aggregated via the same voting strategies to produce a symptom list, which was used for dentists proofreading, thereby facilitating the construction of an intelligent, efficient, and clinically feasible assisted diagnosis workflow for healthcare institutions. 

\textbf{Evaluation details for health management and intelligent diagnosis.} We collected 305 intraoral images from 61 patients captured via edge devices with 5 images per patient. We invited 3 expert dentists with more than 8 years experience to annotate their manifestation on seven oral disease tasks (i.e., caries, periodontal disease, wedge-shaped defects, demineralization, plaque, tooth wear, and calculus). For malocclusion tasks, we selected 69 patients from the test set with 7 modalities with high annotation quality. The performance is evaluated with a diagnosis matching score $S^{\{X\}}=\frac{1}{L\times N}\sum_{l}^{L}\sum_{i}^{N}s_{l,i}^{\{X\}}$, where $\{{X}\}$ represents different voting strategies $mjv$ or $mcv$, $L$ is the number of patients, $N$ is the number of tasks, and $s_{l,i}$ indicates the score of task $i$ for patient $l$ (Supplementary Note 5). 

\textbf{Multi-agent collaborative framework.} Witnessing the outstanding performance of current reasoning models across various real-world scenarios, we sought to explore integrating a reasoning model onto the DentVLM, thereby constructing a multi-agent collaborative framework to provide a more user-friendly interaction experience. For each input image and question, DentVLM first performs reasoning and produces an initial response. This response is then passed to the selected DeepSeek-R1~\cite{deepseekr1} reasoning model, which refines it into a more professional reply by incorporating the historical conversation context, the user's query, and the initial answer (Figure ~\ref{fig:figure5}.d). We dynamically select the language of the prompt for DeepSeek-R1 based on whether the user query is ZH or EN (Extended Table 14).

\textbf{Evaluation details for multi-agent collaborative framework.} To further validate the effectiveness of this framework, we sampled 50 VQA pairs from the clinical study set, consisting of 50 images and 31 tasks covering all modalities. Using this data, we interacted with the developed framework to generate responses spanning multiple rounds of dialogue: 5 samples with one interaction round, 37 samples with two rounds, and 8 samples with three rounds. Three professional dentists with more than 8 years experience were invited to evaluate the interaction records on seven dimensions: correctness, completeness, fairness, faithfulness, acceptability, readability, and coherence, using a score ranging from 1 to 5, with 1 for poor, 2 for fair, 3 for moderate, 4 for good and 5 for excellent. The first five dimensions share similar meanings with those used to evaluate rationales in the clinical study. The readability refers to whether the generated content is easy to read and understand, while coherence assesses the contextual consistency and logical flow throughout the multi-turn dialogues. 

\section{Computing hardware and software}
Codes for the entire work were implemented in Python (v3.10.16). For the two-stage training and various downstream evaluations, we used a server with 8  80 GB powerful GPUs and Ubuntu 20.04 OS. The training process is configured on the 8-GPU single-node by DeepSpeed ZeRO-2 using LLaMA-Factory~\cite{llamafactory} (\href{https://github.com/hiyouga/LLaMA-Factory}{https://github.com/hiyouga/LLaMA-Factory}) with PyTorch (v2.4.0, CUDA 12.4) and Transformers (v4.45.2). The evaluation of DentVLM is deployed on the single GPU with vLLM (v0.6.3.post1) and vLLM-flash-attn (v2.6.1). For proprietary MLLMs, we employed GPT by OpenAI Batch API (\href{https://platform.openai.com}{https://platform.openai.com}), Gemini by Google Developer API (\href{https://ai.google.dev}{https://ai.google.dev}). For Gemma3, InterVL-2.5, LLaVA-OneVision, Llama-3.2, Qwen2-VL and Qwen2.5-VL, we deployed them in consistent ways with DentVLM. For other models, we implemented them by modification based on their official inference ways: DeepSeek-VL2 (\href{https://github.com/deepseek-ai/DeepSeek-VL2}{https://github.com/deepseek-ai/DeepSeek-VL2}), LLaVA-Med (\href{https://github.com/microsoft/LLaVA-Med}{https://github.com/microsoft/LLaVA-Med}), HuatuoGPT-Vision (\href{https://github.com/FreedomIntelligence/HuatuoGPT-Vision}{https://github.com/FreedomIntelligence/HuatuoGPT-Vision}) and RadFM (\href{https://github.com/chaoyi-wu/RadFM}{https://githu\\b.com/chaoyi-wu/RadFM}). Additionally, all web-based platforms developed in this work utilized Flask as the backend framework, MySQL for database management, Docker and Docker Compose for containerization and orchestration, along with Nginx as a reverse proxy server.

\section{Code and data availability}
The datasets used in this study are not publicly available yet due to privacy considerations. Researchers interested in accessing these in-house datasets may request the corresponding author Z.L. (zuozhuliu@intl.zju.edu.cn). All requests will be processed within 4 weeks, as they must be evaluated in accordance with institutional and governmental policies. The code used for implementation and evaluation will be released after the paper is accepted. All relevant details have been thoroughly described in the Methods and Supplemental Note to support reproducibility.

\section{Author contributions}
Z.L., J.H., Bing Fang, J.S. and J.W. conceived and designed the study. Z.L., J.H., Y.F. and H.W. carried out data acquisition. Z.M., X.D., H.Z., Z.L., J.H., H.W., Bin Feng, Jin Li, X.L., Y.W. and L.X. carried out the data preprocessing. Z.M., H.W., Z.L. and X.D. developed the AI model. X.D. and Z.M. developed web application systems. Z.M., Z.L., X.D., Jiaxiang Liu, J.H., H.W., and H.Z. designed and conducted experiments. Z.L., Bing Fang, Y.F., Z.M., X.D., Jin Li and J.H. carried out the clinical deployment. Z.M., J.H., X.D. Z.L., Jiaxiang Liu, H.W., Bin Feng and T.H. carried out the statistical analysis. Z.M., X.D., X.G., Jiaxiang Liu, Z.L. and H.Z. generated figures. Z.M., Z.L., J.H., Jiaxing Liu, J.S., and X.D. wrote the draft. Z.M., Z.L., J.H., Y.F., Bin Feng, T.H., Y.W., H.X., J.X., X.L., J.Z., F.Z., Z.Z., L.X., Bing Fang, J.S., and J.W. revised the paper.

\section{Acknowledgement}
This work is supported by the National Natural Science Foundation of China (Grant No. 12326612, 62476241), the Natural Science Foundation of Zhejiang Province, China (Grant No. LZ23F020008), the Zhejiang University-Angelalign Inc. R\&D Center for Intelligent Healthcare, and the State Key Laboratory (SKL) of Biobased Transportation Fuel Technology.

\clearpage
\bibliographystyle{unsrt}
\bibliography{references}

\begin{thebibliography}{10}

\bibitem{jain2024s}
Nityanand Jain, Upasna Dutt, Igor Radenkov, and Shivani Jain.
\newblock Who's global oral health status report 2022: Actions, discussion and implementation, 2024.

\bibitem{nyamuryekung2015relative}
Kasusu~K Nyamuryekung’e, Satu~M Lahti, and Risto~J Tuominen.
\newblock The relative patient costs and availability of dental services, materials and equipment in public oral care facilities in tanzania.
\newblock {\em BMC oral health}, 15(1):74, 2015.

\bibitem{uguru2020access}
Nkolika Uguru, Obinna Onwujekwe, Udochukwu~Ugochukwu Ogu, and Chibuzo Uguru.
\newblock Access to oral health care: a focus on dental caries treatment provision in enugu nigeria.
\newblock {\em BMC Oral Health}, 20(1):145, 2020.

\bibitem{masood2015household}
Mohd Masood, Aubrey Sheiham, and Eduardo Bernab{\'e}.
\newblock Household expenditure for dental care in low and middle income countries.
\newblock {\em PLoS one}, 10(4):e0123075, 2015.

\bibitem{boutayeb2010burden}
Abdesslam Boutayeb.
\newblock The burden of communicable and non-communicable diseases in developing countries.
\newblock In {\em Handbook of disease burdens and quality of life measures}, pages 531--546. Springer, 2010.

\bibitem{seminario2020mitigating}
Ana~Lucia Seminario, Timothy DeRouen, Mimansa Cholera, Jennifer Liu, Prathip Phantumvanit, Arthur Kemoli, Jorge Castillo, and Waranuch Pitiphat.
\newblock Mitigating global oral health inequalities: Research training programs in low-and middle-income countries.
\newblock {\em Annals of Global Health}, 86(1):141, 2020.

\bibitem{luan2024universal}
Yiqun Luan, Divesh Sardana, Ashiana Jivraj, David Liu, Nishmi Abeyweera, Yajin Zhao, Jacqueline Cellini, Michelle Bass, Jing Wang, Xinran Lu, et~al.
\newblock Universal coverage for oral health care in 27 low-income countries: a scoping review.
\newblock {\em Global Health Research and Policy}, 9(1):34, 2024.

\bibitem{benzian2011political}
Habib Benzian, Martin Hobdell, Christopher Holmgren, Robert Yee, Bella Monse, Johannes~T Barnard, and Wim van Palenstein~Helderman.
\newblock Political priority of global oral health: an analysis of reasons for international neglect.
\newblock {\em International dental journal}, 61(3):124--130, 2011.

\bibitem{watt2019ending}
Richard~G Watt, Bl{\'a}naid Daly, Paul Allison, Lorna~MD Macpherson, Renato Venturelli, Stefan Listl, Robert~J Weyant, Manu~R Mathur, Carol~C Guarnizo-Herre{\~n}o, Roger~Keller Celeste, et~al.
\newblock Ending the neglect of global oral health: time for radical action.
\newblock {\em The Lancet}, 394(10194):261--272, 2019.

\bibitem{rashid2022hybrid}
Umer Rashid, Aiman Javid, Abdur~Rehman Khan, Leo Liu, Adeel Ahmed, Osman Khalid, Khalid Saleem, Shaista Meraj, Uzair Iqbal, and Raheel Nawaz.
\newblock A hybrid mask rcnn-based tool to localize dental cavities from real-time mixed photographic images.
\newblock {\em PeerJ Computer Science}, 8:e888, 2022.

\bibitem{islam2022teledentistry}
Md~Refat~Readul Islam, Rafiqul Islam, Sultana Ferdous, Chiharu Watanabe, Monica Yamauti, Mohammad~Khursheed Alam, and Hidehiko Sano.
\newblock Teledentistry as an effective tool for the communication improvement between dentists and patients: an overview.
\newblock In {\em Healthcare}, volume~10, page 1586. MDPI, 2022.

\bibitem{jang2022accurate}
Woo~Sung Jang, Sunjai Kim, Pill~Sang Yun, Han~Sol Jang, You~Won Seong, Hee~Soo Yang, and Jae-Seung Chang.
\newblock Accurate detection for dental implant and peri-implant tissue by transfer learning of faster r-cnn: a diagnostic accuracy study.
\newblock {\em BMC Oral Health}, 22(1):591, 2022.

\bibitem{cui2019toothnet}
Zhiming Cui, Changjian Li, and Wenping Wang.
\newblock Toothnet: automatic tooth instance segmentation and identification from cone beam ct images.
\newblock In {\em Proceedings of the IEEE/CVF conference on computer vision and pattern recognition}, pages 6368--6377, 2019.

\bibitem{liu2023deep}
Jiaxiang Liu, Jin Hao, Hangzheng Lin, Wei Pan, Jianfei Yang, Yang Feng, Gaoang Wang, Jin Li, Zuolin Jin, Zhihe Zhao, et~al.
\newblock Deep learning-enabled 3d multimodal fusion of cone-beam ct and intraoral mesh scans for clinically applicable tooth-bone reconstruction.
\newblock {\em Patterns}, 4(9), 2023.

\bibitem{hao2022toward}
J~Hao, W~Liao, YL~Zhang, J~Peng, Z~Zhao, Z~Chen, BW~Zhou, Y~Feng, B~Fang, ZZ~Liu, et~al.
\newblock Toward clinically applicable 3-dimensional tooth segmentation via deep learning.
\newblock {\em Journal of dental research}, 101(3):304--311, 2022.

\bibitem{xiong2023tsegformer}
Huimin Xiong, Kunle Li, Kaiyuan Tan, Yang Feng, Joey~Tianyi Zhou, Jin Hao, Haochao Ying, Jian Wu, and Zuozhu Liu.
\newblock Tsegformer: 3d tooth segmentation in intraoral scans with geometry guided transformer.
\newblock In {\em International conference on medical image computing and computer-assisted intervention}, pages 421--432. Springer, 2023.

\bibitem{shi2024leta}
Zefeng Shi, Zijie Meng, Ruizhe Chen, Yang Feng, Zeyu Zhao, Jin Hao, Bing Fang, Zuozhu Liu, and Youyi Zheng.
\newblock Leta: tooth alignment prediction based on dual-branch latent encoding.
\newblock {\em IEEE Transactions on Visualization and Computer Graphics}, 2024.

\bibitem{elgarba2024validation}
Bahaaeldeen~M Elgarba, Rocharles~Cavalcante Fontenele, Saleem Ali, Abdullah Swaity, Jan Meeus, Sohaib Shujaat, and Reinhilde Jacobs.
\newblock Validation of a novel ai-based automated multimodal image registration of cbct and intraoral scan aiding presurgical implant planning.
\newblock {\em Clinical Oral Implants Research}, 35(11):1506--1517, 2024.

\bibitem{abdinian2024comparison}
Mehrdad Abdinian, Maedeh Aminian, Forouzan Keymasi, Parisa Soltani, Mariangela Cernera, Niccolo~Giuseppe Armogida, and Gianrico Spagnuolo.
\newblock Comparison of the operator and surrounding dose when using portable intraoral x-ray devices.
\newblock {\em Applied Sciences}, 14(8):3515, 2024.

\bibitem{park2009portable}
Wonse Park, Dong-Keun Kim, Jung-Chae Kim, Kee-Deog Kim, and Sun~K Yoo.
\newblock A portable dental image viewer using a mobile network to provide a tele-dental service.
\newblock {\em Journal of telemedicine and telecare}, 15(3):145--149, 2009.

\bibitem{glick2021fdi}
Michael Glick and David~M Williams.
\newblock Fdi vision 2030: delivering optimal oral health for all.
\newblock {\em International dental journal}, 71(1):3, 2021.

\bibitem{patel2023chatgpt}
Sajan~B Patel and Kyle Lam.
\newblock Chatgpt: the future of discharge summaries?
\newblock {\em The Lancet Digital Health}, 5(3):e107--e108, 2023.

\bibitem{ali2023using}
Stephen~R Ali, Thomas~D Dobbs, Hayley~A Hutchings, and Iain~S Whitaker.
\newblock Using chatgpt to write patient clinic letters.
\newblock {\em The Lancet Digital Health}, 5(4):e179--e181, 2023.

\bibitem{singhal2023large}
Karan Singhal, Shekoofeh Azizi, Tao Tu, S~Sara Mahdavi, Jason Wei, Hyung~Won Chung, Nathan Scales, Ajay Tanwani, Heather Cole-Lewis, Stephen Pfohl, et~al.
\newblock Large language models encode clinical knowledge.
\newblock {\em Nature}, 620(7972):172--180, 2023.

\bibitem{ayers2023comparing}
John~W Ayers, Adam Poliak, Mark Dredze, Eric~C Leas, Zechariah Zhu, Jessica~B Kelley, Dennis~J Faix, Aaron~M Goodman, Christopher~A Longhurst, Michael Hogarth, et~al.
\newblock Comparing physician and artificial intelligence chatbot responses to patient questions posted to a public social media forum.
\newblock {\em JAMA internal medicine}, 183(6):589--596, 2023.

\bibitem{umer2024innovation}
Fahad Umer, Itrat Batool, and Nighat Naved.
\newblock Innovation and application of large language models (llms) in dentistry--a scoping review.
\newblock {\em BDJ open}, 10(1):90, 2024.

\bibitem{huang2023chatgpt}
Hanyao Huang, Ou~Zheng, Dongdong Wang, Jiayi Yin, Zijin Wang, Shengxuan Ding, Heng Yin, Chuan Xu, Renjie Yang, Qian Zheng, et~al.
\newblock Chatgpt for shaping the future of dentistry: the potential of multi-modal large language model.
\newblock {\em International Journal of Oral Science}, 15(1):29, 2023.

\bibitem{gpt4o}
Aaron Hurst, Adam Lerer, Adam~P Goucher, Adam Perelman, Aditya Ramesh, Aidan Clark, AJ~Ostrow, Akila Welihinda, Alan Hayes, Alec Radford, et~al.
\newblock Gpt-4o system card.
\newblock {\em arXiv preprint arXiv:2410.21276}, 2024.

\bibitem{gemini}
Gemini Team, Rohan Anil, Sebastian Borgeaud, Jean-Baptiste Alayrac, Jiahui Yu, Radu Soricut, Johan Schalkwyk, Andrew~M Dai, Anja Hauth, Katie Millican, et~al.
\newblock Gemini: a family of highly capable multimodal models.
\newblock {\em arXiv preprint arXiv:2312.11805}, 2023.

\bibitem{deepseekvl}
Zhiyu Wu, Xiaokang Chen, Zizheng Pan, Xingchao Liu, Wen Liu, Damai Dai, Huazuo Gao, Yiyang Ma, Chengyue Wu, Bingxuan Wang, et~al.
\newblock Deepseek-vl2: Mixture-of-experts vision-language models for advanced multimodal understanding.
\newblock {\em arXiv preprint arXiv:2412.10302}, 2024.

\bibitem{gemma3}
Gemma Team, Aishwarya Kamath, Johan Ferret, Shreya Pathak, Nino Vieillard, Ramona Merhej, Sarah Perrin, Tatiana Matejovicova, Alexandre Ram{\'e}, Morgane Rivi{\`e}re, et~al.
\newblock Gemma 3 technical report.
\newblock {\em arXiv preprint arXiv:2503.19786}, 2025.

\bibitem{internvl2.5}
Zhe Chen, Weiyun Wang, Yue Cao, Yangzhou Liu, Zhangwei Gao, Erfei Cui, Jinguo Zhu, Shenglong Ye, Hao Tian, Zhaoyang Liu, et~al.
\newblock Expanding performance boundaries of open-source multimodal models with model, data, and test-time scaling.
\newblock {\em arXiv preprint arXiv:2412.05271}, 2024.

\bibitem{llava-onevision}
Bo~Li, Yuanhan Zhang, Dong Guo, Renrui Zhang, Feng Li, Hao Zhang, Kaichen Zhang, Peiyuan Zhang, Yanwei Li, Ziwei Liu, et~al.
\newblock Llava-onevision: Easy visual task transfer.
\newblock {\em arXiv preprint arXiv:2408.03326}, 2024.

\bibitem{llama3}
Abhimanyu Dubey, Abhinav Jauhri, Abhinav Pandey, Abhishek Kadian, Ahmad Al-Dahle, Aiesha Letman, Akhil Mathur, Alan Schelten, Amy Yang, Angela Fan, et~al.
\newblock The llama 3 herd of models.
\newblock {\em arXiv e-prints}, pages arXiv--2407, 2024.

\bibitem{qwen2vl}
Peng Wang, Shuai Bai, Sinan Tan, Shijie Wang, Zhihao Fan, Jinze Bai, Keqin Chen, Xuejing Liu, Jialin Wang, Wenbin Ge, et~al.
\newblock Qwen2-vl: Enhancing vision-language model's perception of the world at any resolution.
\newblock {\em arXiv preprint arXiv:2409.12191}, 2024.

\bibitem{qwen2.5vl}
Shuai Bai, Keqin Chen, Xuejing Liu, Jialin Wang, Wenbin Ge, Sibo Song, Kai Dang, Peng Wang, Shijie Wang, Jun Tang, et~al.
\newblock Qwen2. 5-vl technical report.
\newblock {\em arXiv preprint arXiv:2502.13923}, 2025.

\bibitem{llavamed}
Chunyuan Li, Cliff Wong, Sheng Zhang, Naoto Usuyama, Haotian Liu, Jianwei Yang, Tristan Naumann, Hoifung Poon, and Jianfeng Gao.
\newblock Llava-med: Training a large language-and-vision assistant for biomedicine in one day.
\newblock {\em Advances in Neural Information Processing Systems}, 36:28541--28564, 2023.

\bibitem{huatuogptvision}
Junying Chen, Chi Gui, Ruyi Ouyang, Anningzhe Gao, Shunian Chen, Guiming~Hardy Chen, Xidong Wang, Ruifei Zhang, Zhenyang Cai, Ke~Ji, et~al.
\newblock Huatuogpt-vision, towards injecting medical visual knowledge into multimodal llms at scale.
\newblock {\em arXiv preprint arXiv:2406.19280}, 2024.

\bibitem{radfm}
Chaoyi Wu, Xiaoman Zhang, Ya~Zhang, Hui Hui, Yanfeng Wang, and Weidi Xie.
\newblock Towards generalist foundation model for radiology by leveraging web-scale 2d\&3d medical data.
\newblock {\em Nature Communications}, 16(1):7866, 2025.

\bibitem{lora}
Edward~J. Hu, Yelong Shen, Phillip Wallis, Zeyuan Allen-Zhu, Yuanzhi Li, Shean Wang, Lu~Wang, and Weizhu Chen.
\newblock Lora: Low-rank adaptation of large language models, 2021.

\bibitem{deepseekr1}
Daya Guo, Dejian Yang, Haowei Zhang, Junxiao Song, Ruoyu Zhang, Runxin Xu, Qihao Zhu, Shirong Ma, Peiyi Wang, Xiao Bi, et~al.
\newblock Deepseek-r1: Incentivizing reasoning capability in llms via reinforcement learning.
\newblock {\em arXiv preprint arXiv:2501.12948}, 2025.

\bibitem{qiu2022darch}
Liangdong Qiu, Chongjie Ye, Pei Chen, Yunbi Liu, Xiaoguang Han, and Shuguang Cui.
\newblock Darch: Dental arch prior-assisted 3d tooth instance segmentation.
\newblock {\em arXiv preprint arXiv:2204.11911}, 2022.

\bibitem{cui2021tsegnet}
Zhiming Cui, Changjian Li, Nenglun Chen, Guodong Wei, Runnan Chen, Yuanfeng Zhou, Dinggang Shen, and Wenping Wang.
\newblock Tsegnet: An efficient and accurate tooth segmentation network on 3d dental model.
\newblock {\em Medical Image Analysis}, 69:101949, 2021.

\bibitem{wu2022two}
Tai-Hsien Wu, Chunfeng Lian, Sanghee Lee, Matthew Pastewait, Christian Piers, Jie Liu, Fan Wang, Li~Wang, Chiung-Ying Chiu, Wenchi Wang, et~al.
\newblock Two-stage mesh deep learning for automated tooth segmentation and landmark localization on 3d intraoral scans.
\newblock {\em IEEE transactions on medical imaging}, 41(11):3158--3166, 2022.

\bibitem{chung2020pose}
Minyoung Chung, Minkyung Lee, Jioh Hong, Sanguk Park, Jusang Lee, Jingyu Lee, Il-Hyung Yang, Jeongjin Lee, and Yeong-Gil Shin.
\newblock Pose-aware instance segmentation framework from cone beam ct images for tooth segmentation.
\newblock {\em Computers in Biology and Medicine}, 120:103720, 2020.

\bibitem{jang2021fully}
Tae~Jun Jang, Kang~Cheol Kim, Hyun~Cheol Cho, and Jin~Keun Seo.
\newblock A fully automated method for 3d individual tooth identification and segmentation in dental cbct.
\newblock {\em IEEE transactions on pattern analysis and machine intelligence}, 44(10):6562--6568, 2021.

\bibitem{cui2022fully}
Zhiming Cui, Yu~Fang, Lanzhuju Mei, Bojun Zhang, Bo~Yu, Jiameng Liu, Caiwen Jiang, Yuhang Sun, Lei Ma, Jiawei Huang, et~al.
\newblock A fully automatic ai system for tooth and alveolar bone segmentation from cone-beam ct images.
\newblock {\em Nature communications}, 13(1):2096, 2022.

\bibitem{liu2022hierarchical}
Zuozhu Liu, Xiaoxuan He, Hualiang Wang, Huimin Xiong, Yan Zhang, Gaoang Wang, Jin Hao, Yang Feng, Fudong Zhu, and Haoji Hu.
\newblock Hierarchical self-supervised learning for 3d tooth segmentation in intra-oral mesh scans.
\newblock {\em IEEE Transactions on Medical Imaging}, 42(2):467--480, 2022.

\bibitem{wei2020tanet}
Guodong Wei, Zhiming Cui, Yumeng Liu, Nenglun Chen, Runnan Chen, Guiqing Li, and Wenping Wang.
\newblock Tanet: towards fully automatic tooth arrangement.
\newblock In {\em European conference on computer vision}, pages 481--497. Springer, 2020.

\bibitem{lingchen2020iorthopredictor}
YANG Lingchen, SHI Zefeng, Wu~Yiqian, LI~Xiang, ZHOU Kun, FU~Hongbo, et~al.
\newblock iorthopredictor: model-guided deep prediction of teeth alignment.
\newblock {\em ACM Transactions on Graphics}, 39(6):216, 2020.

\bibitem{li2020malocclusion}
Xiaoshuang Li, Lei Bi, Jinman Kim, Tingyao Li, Peng Li, Ye~Tian, Bin Sheng, and Dagan Feng.
\newblock Malocclusion treatment planning via pointnet based spatial transformation network.
\newblock In {\em International conference on medical image computing and computer-assisted intervention}, pages 105--114. Springer, 2020.

\bibitem{wang2022tooth}
Chen Wang, Guangshun Wei, Guodong Wei, Wenping Wang, and Yuanfeng Zhou.
\newblock Tooth alignment network based on landmark constraints and hierarchical graph structure.
\newblock {\em IEEE Transactions on Visualization and Computer Graphics}, 30(2):1457--1469, 2022.

\bibitem{schwendicke2021deep}
Falk Schwendicke, Akhilanand Chaurasia, Lubaina Arsiwala, Jae-Hong Lee, Karim Elhennawy, Paul-Georg Jost-Brinkmann, Flavio Demarco, and Joachim Krois.
\newblock Deep learning for cephalometric landmark detection: systematic review and meta-analysis.
\newblock {\em Clinical oral investigations}, 25(7):4299--4309, 2021.

\bibitem{zeng2021cascaded}
Minmin Zeng, Zhenlei Yan, Shuai Liu, Yanheng Zhou, and Lixin Qiu.
\newblock Cascaded convolutional networks for automatic cephalometric landmark detection.
\newblock {\em Medical Image Analysis}, 68:101904, 2021.

\bibitem{guo2025towards}
Dongqian Guo, Wencheng Han, Pang Lyu, Yuxi Zhou, and Jianbing Shen.
\newblock Towards better cephalometric landmark detection with diffusion data generation.
\newblock {\em IEEE Transactions on Medical Imaging}, 2025.

\bibitem{zhu2023artificial}
Junhua Zhu, Zhi Chen, Jing Zhao, Yueyuan Yu, Xiaojuan Li, Kangjian Shi, Fan Zhang, Feifei Yu, Keying Shi, Zhe Sun, et~al.
\newblock Artificial intelligence in the diagnosis of dental diseases on panoramic radiographs: a preliminary study.
\newblock {\em BMC Oral Health}, 23(1):358, 2023.

\bibitem{chen2024cariesxrays}
Bingzhi Chen, Sisi Fu, Yishu Liu, Jiahui Pan, Guangming Lu, and Zheng Zhang.
\newblock Cariesxrays: Enhancing caries detection in hospital-scale panoramic dental x-rays via feature pyramid contrastive learning.
\newblock In {\em Proceedings of the AAAI Conference on Artificial Intelligence}, volume~38, pages 21940--21948, 2024.

\bibitem{hamamci2023diffusion}
Ibrahim~Ethem Hamamci, Sezgin Er, Enis Simsar, Anjany Sekuboyina, Mustafa Gundogar, Bernd Stadlinger, Albert Mehl, and Bjoern Menze.
\newblock Diffusion-based hierarchical multi-label object detection to analyze panoramic dental x-rays.
\newblock In {\em International conference on medical image computing and computer-assisted intervention}, pages 389--399. Springer, 2023.

\bibitem{yang2022histopathology}
SY~Yang, SH~Li, JL~Liu, XQ~Sun, YY~Cen, RY~Ren, SC~Ying, Y~Chen, ZH~Zhao, and W~Liao.
\newblock Histopathology-based diagnosis of oral squamous cell carcinoma using deep learning.
\newblock {\em Journal of Dental Research}, 101(11):1321--1327, 2022.

\bibitem{llava}
Haotian Liu, Chunyuan Li, Qingyang Wu, and Yong~Jae Lee.
\newblock Visual instruction tuning.
\newblock {\em Advances in neural information processing systems}, 36:34892--34916, 2023.

\bibitem{vit}
Alexey Dosovitskiy, Lucas Beyer, Alexander Kolesnikov, Dirk Weissenborn, Xiaohua Zhai, Thomas Unterthiner, Mostafa Dehghani, Matthias Minderer, Georg Heigold, Sylvain Gelly, et~al.
\newblock An image is worth 16x16 words: Transformers for image recognition at scale.
\newblock {\em arXiv preprint arXiv:2010.11929}, 2020.

\bibitem{dehghani2023patch}
Mostafa Dehghani, Basil Mustafa, Josip Djolonga, Jonathan Heek, Matthias Minderer, Mathilde Caron, Andreas Steiner, Joan Puigcerver, Robert Geirhos, Ibrahim~M Alabdulmohsin, et~al.
\newblock Patch n’pack: Navit, a vision transformer for any aspect ratio and resolution.
\newblock {\em Advances in Neural Information Processing Systems}, 36:2252--2274, 2023.

\bibitem{qwen2}
Qwen Team.
\newblock Qwen2 technical report.
\newblock {\em arXiv preprint arXiv:2407.10671}, 2, 2024.

\bibitem{deepspeed}
Jeff Rasley, Samyam Rajbhandari, Olatunji Ruwase, and Yuxiong He.
\newblock Deepspeed: System optimizations enable training deep learning models with over 100 billion parameters.
\newblock In {\em Proceedings of the 26th ACM SIGKDD international conference on knowledge discovery \& data mining}, pages 3505--3506, 2020.

\bibitem{vllm}
Woosuk Kwon, Zhuohan Li, Siyuan Zhuang, Ying Sheng, Lianmin Zheng, Cody~Hao Yu, Joseph Gonzalez, Hao Zhang, and Ion Stoica.
\newblock Efficient memory management for large language model serving with pagedattention.
\newblock In {\em Proceedings of the 29th symposium on operating systems principles}, pages 611--626, 2023.

\bibitem{sigmoid}
Xiaohua Zhai, Basil Mustafa, Alexander Kolesnikov, and Lucas Beyer.
\newblock Sigmoid loss for language image pre-training.
\newblock In {\em Proceedings of the IEEE/CVF international conference on computer vision}, pages 11975--11986, 2023.

\bibitem{llamafactory}
Yaowei Zheng, Richong Zhang, Junhao Zhang, Yanhan Ye, Zheyan Luo, Zhangchi Feng, and Yongqiang Ma.
\newblock Llamafactory: Unified efficient fine-tuning of 100+ language models.
\newblock {\em arXiv preprint arXiv:2403.13372}, 2024.

\end{thebibliography}

\clearpage
\section{Supplementary}
Due to limited space, we have not included the full supplementary materials in this version. All supplementary content is available upon request. Please contact the corresponding author if you are interested in accessing the full materials.

\end{document}